\documentclass[journal]{IEEEtran}
\usepackage{cite}

\ifCLASSINFOpdf
\else
\fi
\usepackage{amsmath,amssymb,bm,graphicx,color}
\newcommand{\mbf}[1]{\mathbf{#1}}
\usepackage[table]{xcolor}
\definecolor{mygray}{gray}{.9}

\allowdisplaybreaks[4]
\usepackage{url}
\usepackage[colorlinks, linkcolor=blue, anchorcolor=blue, citecolor=blue]{hyperref}

\usepackage{amsthm}
\theoremstyle{remark}

\theoremstyle{plain}
\newtheorem{thm}{Theorem}
\newtheorem{definition}{Definition}
\newtheorem{prop}[thm]{Proposition}

\begin{document}

\title{Deep Latent-Variable Kernel Learning}

%

\author{Haitao~Liu,
	Yew-Soon~Ong,~\IEEEmembership{Fellow,~IEEE,}
	Xiaomo~Jiang
	and~Xiaofang~Wang
	\thanks{Haitao Liu and Xiaofang Wang are with School of Energy and Power Engineering, Dalian University of Technology, China, 116024. E-mail: htliu@dlut.edu.cn, dlwxf@dlut.edu.cn.}
	\thanks{Yew-Soon Ong is with School of Computer Science and Engineering, Nanyang Technological University, Singapore, 639798. E-mail: asysong@ntu.edu.sg.}
	\thanks{Xiaomo Jiang is with the Digital Twin Laboratory for Industrial Equipment at Dalian University of Technology, China, 116024. E-mail: xiaomojiang2019@dlut.edu.cn.}
	}

\markboth{}%
{Shell \MakeLowercase{\textit{et al.}}: Bare Demo of IEEEtran.cls for IEEE Journals}

\maketitle

\begin{abstract}
Deep kernel learning (DKL) leverages the connection between Gaussian process (GP) and neural networks (NN) to build an end-to-end, hybrid model. It combines the capability of NN to learn rich representations under massive data and the non-parametric property of GP to achieve automatic regularization that incorporates a trade-off between model fit and model complexity. However, the deterministic encoder may weaken the model regularization of the following GP part, especially on small datasets, due to the free latent representation. We therefore present a complete deep latent-variable kernel learning (DLVKL) model wherein the latent variables perform stochastic encoding for regularized representation. We further enhance the DLVKL from two aspects: (i) the expressive variational posterior through neural stochastic differential equation (NSDE) to improve the approximation quality, and (ii) the hybrid prior taking knowledge from both the SDE prior and the posterior to arrive at a flexible trade-off. Intensive experiments imply that the DLVKL-NSDE performs similarly to the well calibrated GP on small datasets, and outperforms existing deep GPs on large datasets.
\end{abstract}


\section{Introduction}
In the machine learning community, Gaussian process (GP)~\cite{williams2006gaussian} is a well-known Bayesian model to learn the underlying function $f: \mbf{x} \mapsto \mbf{y}$. In comparison to the deterministic, parametric machine learning models, e.g., neural networks (NN), the non-parametric GP could encode user's prior knowledge, calibrate the model complexity automatically, and quantify the uncertainty of prediction, thus showing high flexibility and interpretability. Hence, it has been popularized within various scenarios like regression, classification~\cite{wang2014spectrum}, clustering~\cite{li2020shared}, representation learning~\cite{lawrence2005probabilistic}, sequence learning~\cite{frigola2014variational}, multi-task learning~\cite{liu2018remarks}, active learning and optimization~\cite{shahriari2016taking, luo2019evolutionary}. 

However, the two main weaknesses of GP are its poor scalability and the limited model capability on massive data. Firstly, as an representative of kernel method, the GP employs the kernel function $k(.,.)$ to encode the spatial correlations of $n$ training points into the stochastic process. Consequently, it performs operations on the full-rank kernel matrix $\mbf{K}_{nn} \in R^{n \times n}$, thus raising a cubic time complexity $\mathcal{O}(n^3)$ which prohibits the application in the era of big data. To improve the scalability, various scalable GPs have then been presented and studied. For example, the sparse approximations introduce $m$ ($m \ll n$) inducing variables $\mbf{u}$ to distillate the latent function values $\mbf{f}$ through prior or posterior approximation~\cite{snelson2006sparse, titsias2009variational}, thus reducing the time complexity to $\mathcal{O}(nm^2)$. The variational inference with reorganized evidence lower bound (ELBO) could further make the stochastic gradient descent optimizer, e.g., Adam~\cite{kingma2014adam}, available for training with a greatly reduce complexity of $\mathcal{O}(m^3)$~\cite{hensman2013gaussian}. Moreover, further complexity reduction can be achieved by exploiting the structured inducing points and the iterative methods through matrix-vector multiplies, see for example~\cite{wilson2015kernel, gardner2018product}. In contrast to the global sparse approximation, the complexity of GP can also be reduced through distributed computation~\cite{gal2014distributed, peng2017asynchronous} and local approximation~\cite{deisenroth2015distributed, liu2018generalized}. The idea of divide-and-conquer splits the data for subspace learning, which alleviates the computational burden and helps capturing local patterns. The readers can refer to a recent review~\cite{liu2020gaussian} of scalable GPs for further information.

Secondly, the GP usually uses (i) the Gaussian assumption to have closed-form inference, and (ii) the stationary and smoothing kernels to simply quantify how quickly the correlations vary along dimensions, which thus raise urgent demand for developing new GP paradigms to learn rich statistical representations under massive data. Hence, the interpretation of NN from kernel learning~\cite{lee2017deep} inspires the construction of deep kernels for GP to mimic the nonlinearity and recurrency behaviors of NN~\cite{cho2009kernel, hermans2012recurrent}. But the representation learning of deep kernels in comparison to deep models reviewed below is limited unless they are richly parameterized~\cite{duvenaud2014avoiding}. 

Considering the theoretical connection between GP and wide deep neural networks~\cite{neal1996bayesian, matthews2018gaussian}, a hybrid, end-to-end model, called deep kernel learning (DKL)~\cite{wilson2016deep, wilson2016stochastic, tran2019calibrating}, has been proposed to combine the non-parametric property of GP and the inductive biases of NN. In this framework, the NN plays as a \textit{deterministic} encoder for representation learning, and the sparse GP is built on the latent inputs for providing Bayesian estimations. The NN+GP structure thereafter has been extended for handling semi-supervised learning~\cite{jean2018semi} and time-series forecasting~\cite{al2017learning}. The automatic regularization through the marginal likelihood of the last GP layer is expected to improve the performance of DKL and reduces the requirement of fine-tuning and regularization.  But we find that the deterministic encoder may deteriorate the regularization of DKL, \textit{especially on small datasets}, which will be elaborated in the following sections. Alternatively, we could stack the sparse GPs together to build the deep GP (DGP)~\cite{damianou2013deep, salimbeni2017doubly}, which admits the layer-by-layer GP transformation that yields non-Gaussian distributions. Hence, the DGPs usually resort to the variational inference for model training. Different from DKL, the DGPs employ the full GP paradigm to arrive at automatic regularization. But the representation learning through layer-by-layer sparse GPs suffers from (i) high time complexity, (ii) complicated approximate inference, and (iii) limited capability due to the finite global inducing variables in each layer. 

From the foregoing review and discussion, it is observed that the simple and scalable DKL enjoys great representation power of NN but suffers from the mismatch between the \textit{deterministic} representation and the \textit{stochastic} inference, which may risk over-fitting, especially on small datasets. While the sparse DGP enjoys the well calibrated GP paradigm but suffers from high complexity and limited representation capability.

Therefore, this article presents a complete Bayesian version of DKL which inherits (i) the scalability and representation of DKL and (ii) the regularization of DGP. The main contributions of this article are three-fold:
\begin{itemize}
\item We propose an end-to-end latent-variable framework called deep latent-variable kernel learning (DLVKL). It incorporates a stochastic encoding process for regularized representation learning and a sparse GP part for guarding against over-fitting. The whole stochastic framework ensures that it can fully benefit from the automatic regularization of GP;
\item We further improve the DLVKL by constructing (i) the informative variational posterior rather than the simple Gaussian through neural stochastic differential equations (NSDE) to reduce the gap to exact posterior, and (ii) the flexible prior incorporating the knowledge from both the SDE prior and the variational posterior to arrive at a trade-off. The NSDE transformation improves the representation learning and the trade-off provides an adjustable regularization in various scenarios;
\item We showcase the superiority of DLVKL-NSDE against existing deep GPs through intensive (un)supervised learning tasks. The tensorflow implementations are available at \url{https://github.com/LiuHaiTao01/DLVKL}.
\end{itemize}

The remainder of the article is organized as follows. Section~\ref{sec_scalable_gp} briefly introduces the sparse GPs. Section~\ref{sec_DLVKL} then presents the framework of DLVKL. Thereafter, Section~\ref{sec_DLVKL_NSDE} proposes an enhanced version, named DLVKL-NSDE, through informative posterior and flexible prior. Then extensive numerical experiments are conducted in Section~\ref{sec_exp} to verify the superiority of DLVKL-NSDE on (un)supervised learning tasks. Finally, Section~\ref{sec_con} offers the concluding remarks.

\section{Scalable GPs revisited} \label{sec_scalable_gp}
Let $\mbf{x} = \{\mbf{x}_i \in R^{d_{\mbf{x}}} \}_{i=1}^n = \{\mbf{x}_d \in R^{n} \}_{d=1}^{d_{\mbf{x}}}$ be the collection of $n$ points in the input space $\mathcal{X} \in R^{d_{\mbf{x}}}$, and $\mbf{y} = \{\mbf{y}_i \in R^{d_{\mbf{y}}} \}_{i=1}^n = \{\mbf{y}_d \in R^{n} \}_{d=1}^{d_{\mbf{y}}}$ the observations in the output space $\mathcal{Y} \in R^{d_{\mbf{y}}}$, we seek to infer the latent mappings $\{f_d: R^{d_{\mbf{x}}} \mapsto R \}_{d=1}^{d_{\mbf{y}}}$ from data $\mathcal{D} = \{\mbf{x}, \mbf{y} \}$. To this end, the GP characterizes the distributions of latent functions by placing \textit{independent} zero-mean GP priors as $f_d(\mbf{x}) \sim \mathcal{GP}(0, k_d(\mbf{x}, \mbf{x}'))$, $0 \le d \le d_{\mbf{y}}$. For regression and binary classification, we usually have $d_{\mbf{y}} = 1$; while for multi-class classification and unsupervised learning, we are often facing $d_{\mbf{y}} > 1$.

We are interested in two statistics in the GP paradigm. The first is the marginal likelihood $p(\mbf{y}_d|\mbf{x}) = \int p(\mbf{y}_d|\mbf{f}_d) p(\mbf{f}_d|\mbf{x}) d\mbf{f}_d$, where $p(\mbf{f}_d|\mbf{x}) = \mathcal{N}(\mbf{f}_d|\mbf{0}, \mbf{K}_{nn})$ with $\mbf{K}_{nn} = k(\mbf{x}, \mbf{x}) \in R^{n \times n}$.\footnote{For the sake of simplicity, we here use the same kernel for $d_{\mbf{y}}$ outputs.}  The marginal likelihood $p(\mbf{y}|\mbf{x}) = \prod_{d=1}^{d_{\mbf{y}}} p(\mbf{y}_d|\mbf{x})$, the maximization of which optimizes the hyperparameters, automatically achieves a trade-off between model fit and
model complexity~\cite{williams2006gaussian}. As for $p(\mbf{y}_d|\mbf{f}_d)$, we adopt the Gaussian likelihood $p(\mbf{y}_d|\mbf{f}_d) = \mathcal{N}(\mbf{y}_d|\mbf{f}_d, \nu_d^{\epsilon}\mbf{I})$ for regression and unsupervised learning with continuous outputs given the \textit{i.i.d} noise $\epsilon_d \sim \mathcal{N}(0, \nu_d^{\epsilon})$~\cite{williams2006gaussian, lawrence2005probabilistic}. While for binary classification with discrete outputs $y \in \{0, 1\}$ and multi-class classification with $y \in \{1, \cdots, d_{\mbf{y}}\}$, we have
$p(\mbf{y}_d|\mbf{f}_d) = \mathrm{Benoulli}(\pi(\mbf{f}_d))$ and $p(\mbf{y}_d|\mbf{f}_d) = \mathrm{Categorial}(\pi(\mbf{f}_d))$, respectively, wherein $\pi(.) \in [0,1]$ is an inverse link function that squashes $f$ into the class probability space~\cite{kim2006bayesian}.

The second interested statistic is the posterior $p(\mbf{f}_d|\mbf{y}_d, \mbf{x}) \propto p(\mbf{y}_d|\mbf{f}_d) p(\mbf{f}_d|\mbf{x})$ used to perform prediction at a test point $\mbf{x}_*$ as $p(\mbf{f}_{*}|\mbf{y}, \mbf{x}, \mbf{x}_*) = \int p(\mbf{f}_{*}|\mbf{f}, \mbf{x}, \mbf{x}_*) p(\mbf{f}|\mbf{y}, \mbf{x}) d\mbf{f}$. Note that for non-Gaussian likelihoods, the posterior is intractable and we resort to approximate inference algorithms~\cite{nickisch2008approximations}. 

The scalability of GPs however is severely limited on massive data, since the inversion and determinant of $\mbf{K}_{nn}$ incur $\mathcal{O}(n^3)$ operations. Hence, the sparse approximation~\cite{liu2020gaussian} employs a set of inducing variables $\mbf{u}_d \in R^{m}$ with $m \ll n$ at $\widetilde{\mbf{x}} \in R^{m \times d_{\mbf{x}}}$ to distillate the latent function values $\mbf{f}_d$. Then, we have
\begin{align*}
\begin{split}
p(\mbf{u}_d) &= \mathcal{N}(\mbf{u}_d|\mbf{0}, \mbf{K}_{mm}),\\
p(\mbf{f}_d|\mbf{u}_d, \mbf{x}) &= \mathcal{N}(\mbf{f}_d|\mbf{K}_{nm} \mbf{K}_{mm}^{-1} \mbf{u}_d, \mbf{K}_{nn} - \mbf{K}_{nm} \mbf{K}_{mm}^{-1} \mbf{K}_{nm}^{\mathsf{T}}),
\end{split}
\end{align*}
where $\mbf{K}_{mm} = k(\widetilde{\mbf{x}}, \widetilde{\mbf{x}})$ and $\mbf{K}_{nm} = k(\mbf{x}, \widetilde{\mbf{x}})$. Thereafter, variational inference could help handle the intractable $\log p(\mbf{y}|\mbf{x})$. This is conducted by using a tractable variational posterior $q(\mbf{u}_d) = \mathcal{N}(\mbf{u}_d|\mbf{m}_d, \mbf{S}_d)$, and maximizing the KL divergence
\begin{align*}
\mathrm{KL}[q(\mbf{f}, \mbf{u}|\mbf{x}) || p(\mbf{f}, \mbf{u}|\mbf{x}, \mbf{y})] = \sum_{d=1}^{d_{\mbf{y}}} \mathrm{KL}[q(\mbf{f}_d, \mbf{u}_d|\mbf{x}) || p(\mbf{f}_d, \mbf{u}_d|\mbf{x}, \mbf{y}_d)],
\end{align*}
where $p(\mbf{f}_d, \mbf{u}_d| \mbf{x}, \mbf{y}_d) = p(\mbf{f}_d|\mbf{u}_d, \mbf{x}) p(\mbf{u}_d|\mbf{y}_d)$ and $q(\mbf{f}_d, \mbf{u}_d| \mbf{x}) = p(\mbf{f}_d|\mbf{u}_d, \mbf{x}) q(\mbf{u}_d)$. It is equivalent to maximizing the following ELBO
\begin{align} \label{eq_elbo_gp}
\mathcal{L}_{\mathrm{gp}} = \mathbb{E}_{q(\mbf{f}|\mbf{x})} [\log p(\mbf{y}|\mbf{f})] - \mathrm{KL}[q(\mbf{u}) || p(\mbf{u})].
\end{align}
The likelihood term of $\mathcal{L}_{\mathrm{gp}}$ represents the fitting error, and it factorizes over both data points and dimensions
\begin{align*}
\begin{split}
\mathbb{E}_{q(\mbf{f}|\mbf{x})} [\log p(\mbf{y}|\mbf{f})] &= \sum_{d=1}^{d_{\mbf{y}}} \mathbb{E}_{q(\mbf{f}_d|\mbf{x})} [\log p(\mbf{y}_d|\mbf{f}_d)] \\
&= \sum_{i=1}^{n} \mathbb{E}_{q(\mbf{f}_i|\mbf{x}_i)} [\log p(\mbf{y}_i|\mbf{f}_i)],
\end{split}
\end{align*}
thus having a remarkably reduced time complexity of $\mathcal{O}(m^3)$ when performing stochastic optimization. Note that $q(\mbf{f}_d|\mbf{x}) = \int p(\mbf{f}_d|\mbf{u}_d, \mbf{x}) q(\mbf{u}_d) d\mbf{u}_d = \mathcal{N}(\mbf{f}_d|\bm{\mu}^f_d, \bm{\Sigma}_d^f)$ is conditioned on the whole training points, where $\bm{\mu}^f_d = \mbf{K}_{nm}\mbf{K}_{mm}^{-1}\mbf{m}_d$ and $\bm{\Sigma}_d^f = \mbf{K}_{nn}-\mbf{K}_{nm}\mbf{K}_{mm}^{-1}[\mbf{I}-\mbf{S}_d\mbf{K}_{mm}^{-1}]\mbf{K}_{nm}^{\mathsf{T}}$. While the posterior $q(\mbf{f}_i|\mbf{x}_i) = \prod_{d=1}^{d_{\mbf{y}}} \int p(f_{id}|\mbf{u}_d, \mbf{x}_i) q(\mbf{u}_d) d\mbf{u}_d = \mathcal{N}(\mbf{f}_i|\bm{\mu}^f_i, \bm{\nu}^f_i)$ only depends on the related point $\mbf{x}_i$, where $\bm{\mu}^f_i \in R^{d_{\mbf{y}}}$ collects the $i$-th element from each in the set $\{\bm{\mu}^f_d \}_{d=1}^{d_{\mbf{y}}}$; and $\bm{\nu}^f_i \in R^{d_{\mbf{y}}}$ collects the $i$-th diagonal element from each in the set $\{\bm{\Sigma}^f_d \}_{d=1}^{d_{\mbf{y}}}$.\footnote{This viewpoint inspires the doubly stochastic variational inference~\cite{salimbeni2017doubly}. Besides, the covariance here is summarized by the inducing variables.} Besides, the analytical KL term in~\eqref{eq_elbo_gp} guards against over-fitting and seeks to deliver a good inducing set. Note that the likelihood $p(\mbf{y}|\mbf{f})$ in~\eqref{eq_elbo_gp} is not limited to Gaussian. By carefully reorganizing the formulations, we could derive analytical expressions for $\mathcal{L}_{\mathrm{gp}}$ for (un)supervised learning tasks~\cite{hensman2013gaussian, damianou2016variational, liu2019scalable}.

Finally, when the inputs $\mbf{x}$ are unobservable, we use the unsupervised ELBO for $\log p(\mbf{y})$ as
\begin{align}
\begin{split}
\mathcal{L}_{\mathrm{gp}}^{\mathrm{u}} = &\mathbb{E}_{q(\mbf{f}|\mbf{x}) q(\mbf{x})} [\log p(\mbf{y}|\mbf{f})] - \mathrm{KL}[q(\mbf{x}) || p(\mbf{x})] \\
&- \mathrm{KL}[q(\mbf{u}) || p(\mbf{u})]
\end{split}
\end{align}
to infer the latent variables $\mbf{x}$ under the GP latent variable (GPLVM) framework~\cite{titsias2010bayesian}. This unsupervised model can be used for dimensionality reduction, data imputation and density estimation~\cite{damianou2016variational}.

\section{Deep latent-variable kernel learning} \label{sec_DLVKL}
We consider a GP with additional $d_{\mbf{z}}$-dimensional latent variables $\mbf{z}$ as\footnote{We could describe most of deep GPs by the model~\eqref{eq_dlvkl_model}, see Appendix~\ref{app_dgps}.}
\begin{align} \label{eq_dlvkl_model}
p(\mbf{y}|\mbf{x}) = \int p(\mbf{y}|\mbf{z}) p(\mbf{z}|\mbf{x}) d\mbf{z},
\end{align}
where the conditional distribution $p(\mbf{z}|\mbf{x})$ indicates a stochastic encoding of the original input $\mbf{x}$, which could ease the inference of the following generative model $p(\mbf{y}|\mbf{z})$. As a result, the log marginal likelihood satisfies
\begin{align} \label{eq_VAE_elbo}
\log p(\mbf{y}|\mbf{x}) \ge \mathbb{E}_{ q(\mbf{z}|\mbf{x})} [\log p(\mbf{y}|\mbf{z})] - \mathrm{KL}[q(\mbf{z}|\mbf{x}) || p(\mbf{z}|\mbf{x})].
\end{align}
In~\eqref{eq_VAE_elbo}, 
\begin{itemize}
\item as for $p(\mbf{y}|\mbf{z}) = \int p(\mbf{y}|\mbf{f}) p(\mbf{f}|\mbf{z}) d\mbf{f}$, we use independent GPs $f_d \sim \mathcal{GP}(0, k_d(.,.))$, $0 \le d \le d_{\mbf{y}}$, to fit the mappings between $\mbf{y}$ and $\mbf{z}$. Consequently, we have the following lower bound by resorting to the sparse GP as
\begin{align} \label{eq_gp_elbo}
\log p(\mbf{y}|\mbf{z}) \ge \mathbb{E}_{q(\mbf{f}|\mbf{u}, \mbf{z}) q(\mbf{u})} [\log p(\mbf{y}|\mbf{f})] - \mathrm{KL}[q(\mbf{u}) || p(\mbf{u})].
\end{align}
Note that the GP mapping employed here learns a joint distribution $p(\mbf{y}_d|\mbf{z}) = \mathcal{N}(\mbf{y}_d|\mbf{0}, \mbf{K}_{nn} + \nu_d^{\epsilon} \mbf{I})$ by considering the correlations over the entire space in order to achieve automatic regularization;
\item as for the prior $p(\mbf{z}|\mbf{x})$, we usually employ the fully independent form $p(\mbf{z}|\mbf{x}) = \prod_{i=1}^n \mathcal{N}(\mbf{z}_i|\mbf{x}_i) = \prod_{i=1}^n \prod_{d=1}^{d_{\mbf{z}}} \mathcal{N}(z_{i,d}|\mbf{x}_i)$ factorized over both data index $i$ and dimension index $d$;
\item  as for the decoder $q(\mbf{z}|\mbf{x})$, since we aim to utilize the great representational power of NN, it often takes the independent Gaussians $q(\mbf{z}|\mbf{x}) = \prod_{i=1}^n \mathcal{N}(\mbf{z}_i|\bm{\mu}_i, \mathrm{diag}[\bm{\nu}_i])$. Instead of directly treating the means $\{\bm{\mu}_i\}_{i=1}^n$ and variances $\{\bm{\nu}_i\}_{i=1}^n$ as hyperparameters, they are made of parameterized function of the input $\mbf{x}_i$ through multi-layer perception (MLP), for example, as
\begin{align*}
\bm{\mu}_i = \mathtt{Linear}(\mathtt{MLP}(\mbf{x}_i)), \quad
\bm{\nu}_i = \mathtt{Softplus}(\mathtt{MLP}(\mbf{x}_i)).
\end{align*}
This strategy is called amortized variational inference that shares the parameters over all training points, thus allowing for efficient training. 
\end{itemize}

Combing~\eqref{eq_VAE_elbo} and~\eqref{eq_gp_elbo}, the final ELBO for the proposed deep latent-variable kernel learning (DLVKL) writes as
\begin{align} \label{eq_gpvae_elbo}
\begin{split}
\mathcal{L} = &\mathbb{E}_{q(\mbf{f}|\mbf{z}) q(\mbf{z}|\mbf{x})} [\log p(\mbf{y}|\mbf{f})] - \mathrm{KL}[q(\mbf{z}|\mbf{x}) || p(\mbf{z}|\mbf{x})] \\
&- \mathrm{KL}[q(\mbf{u}) || p(\mbf{u})],
\end{split}
\end{align}
which can be optimized by the reparameterization trick~\cite{kingma2013auto}. 

It is found that in comparison to the ELBO of DKL in Appendix~\ref{app_dkl}, the additional $\mathrm{KL}[q(\mbf{z}|\mbf{x}) || p(\mbf{z}|\mbf{x})]$ in~\eqref{eq_gpvae_elbo} \textit{regularizes the representation learning of encoder}, which distinguishes our DLVKL from the DKL proposed in~\cite{wilson2016deep}. The DKL adopts a \textit{deterministic} representation learning $\mbf{z} = \mathtt{MLP}(\mbf{x})$, which is equivalent to the variational distribution $q(\mbf{z}_i|\mbf{x}_i) = \mathcal{N}(\mbf{z}_i|\bm{\mu}_i, \bm{\nu}_i \rightarrow \mbf{0}^{+}) = \delta(\mbf{z}_i - \bm{\mu}_i)$. Intuitively, in order to maximize $\mathcal{L}$, it is encouraged to learn a $p(\mbf{y}|\mbf{z})$ which maps $\mbf{z}$ to a distribution concentrated on $\mbf{y}$. Particularly, pushing all the mass of the distribution $p(\mbf{y}|\mbf{z})$ on $\mbf{y}$ results in $\mathcal{L} \rightarrow \infty$, which however risks severe over-fitting~\cite{bowman2015generating}. The last GP layer employed in DKL is expected to alleviate this issue by considering the joint correlations over the output space. But the deterministic encoder, which is inconsistent to the following stochastic GP part, will incur free latent representation to weaken the model regularization, which results in over-fitting and under-estimated prediction variance in scenarios with finite data points. Contrarily, the proposed DLVKL builds a complete statistical learning framework wherein the additional KL regularization could avoid the issue by pushing $q(\mbf{z}|\mbf{x})$ to match the prior $p(\mbf{z}|\mbf{x})$.

Note that when $\mbf{x}$ is unobservable, i.e., $\mbf{x} \triangleq \mbf{y}$, the proposed DLVKL recovers the GPLVM using back constraints (recognition model)~\cite{bui2015stochastic}. Also, the bound~\eqref{eq_gpvae_elbo} becomes the VAE-type ELBO
\begin{align} \label{eq_VAE_unsupervised_elbo}
\begin{split}
\mathcal{L}^{\mathrm{u}} = &\mathbb{E}_{q(\mbf{f}|\mbf{z}) q(\mbf{z})} [\log p(\mbf{y}|\mbf{f})] - \mathrm{KL}[q(\mbf{z}) || p(\mbf{z})] \\
&- \mathrm{KL}[q(\mbf{u}) || p(\mbf{u})]
\end{split}
\end{align}
for unsupervised learning~\cite{kingma2013auto}, wherein the main difference is that a GP decoder is employed.

Though the complete stochastic framework makes DLVKL be more attractive than DKL, it has two challenges to be addressed: 
\begin{itemize}
\item firstly, the assumed variational Gaussian posterior $q(\mbf{z}|\mbf{x})$ is often significantly different from the exact posterior $p(\mbf{z}|\mbf{x}, \mbf{y})$. This gap may deteriorate the performance and thus raises the demand of expressive  $q(\mbf{z}|\mbf{x})$ for better approximation, which will be addressed in Section~\ref{sec_info_post};
\item secondly, the choice of prior $p(\mbf{z}|\mbf{x})$ affects the regularization $\mathrm{KL}[q(\mbf{z}|\mbf{x}) || p(\mbf{z}|\mbf{x})]$ on latent representation. The flexibility and expressivity of the prior will be improved in Section~\ref{sec_flexible_prior}.
\end{itemize}

\section{Improvements of DLVKL} \label{sec_DLVKL_NSDE}
The improvements of DLVKL come from two aspects: (i) the more expressive variational posterior transformed through neural stochastic differential equation (NSDE) for better approximating the exact posterior; and (ii) the flexible, hybrid prior to introduce adjustable regularization on  latent representation. The two improvements are elaborated respectively in the following subsections.

\subsection{Expressive variational posterior $q(\mbf{z}|\mbf{x})$ via NSDE} \label{sec_info_post}
In order to generate expressive posterior $q(\mbf{z}|\mbf{x})$ rather than the simple Gaussian, which is beneficial for minimizing $\mathrm{KL}[q(\mbf{z}|\mbf{x}) || p(\mbf{z}|\mbf{x}, \mbf{y})]$, we interpret the stochastic encoding $\mbf{x} \rightarrow \mbf{z}$ as a \textit{continuous-time} dynamic system governed by SDE~\cite{friedrich2011approaching} over the time period $[0, T]$ as
\begin{align} \label{eq_nsde_continuous}
d\mbf{z}_i^t = \bm{\mu}_i^t + \mbf{L}_i^t d\mbf{w}^t, \quad 0 \le t \le T,
\end{align}
where the initial state $\mbf{z}_i^0 \triangleq \mbf{x}_i$; $\bm{\mu}_i^t = \mu(\mbf{z}_i^t)$ is the deterministic drift vector; $\bm{\Sigma}_i^t = \mbf{L}_i^t (\mbf{L}_i^t)^{\mathsf{T}} = \mathrm{diag}[\bm{\nu}_i^t]$ with $\mbf{L}_i^t = L(\mbf{z}_i^t)$ is the positive definite diffusion matrix which indicates the scale of the random Brownian motion $\mbf{w}^t$ that scatters the state with random perturbation; and $\mbf{w}^t$ represents the standard and uncorrelated Brownian process, which starts from a zero initial state $\mbf{w}^0 = \mbf{0}$ and has independent Gaussian increment $\mbf{w}^{t+\Delta t} - \mbf{w}^t \sim \mathcal{N}(\mbf{0}, \Delta t\mbf{I})$. 

The SDE flow in~\eqref{eq_nsde_continuous} defines a sequence of transformation indexed on a continuous-time domain, the purpose of which is to evolve the simple initial state to the one with expressive distribution. In comparison to the normalizing flow~\cite{rezende2015variational} indexed on a discrete-time domain, the SDE flow is more theoretically grounded since it could approach any distribution asymptotically~\cite{chen2018continuous}. Besides, the diffusion term, which distinguishes SDE from the ordinary differential equation (ODE)~\cite{chen2018neural}, makes the flow more stable and robust from the view of regularization~\cite{liu2019neural}.

The solution of SDE is given by the It\^{o} integral which integrates the state from the initial state to time $t$ as
\begin{align*}
\mbf{z}_i^t = \mbf{z}_i^0 + \int_0^t \bm{\mu}_i^{\tau} d\tau + \int_0^t \mbf{L}_i^{\tau} d\mbf{w}^{\tau}.
\end{align*}
Note that due to the non-differentiable $\mbf{w}^{\tau}$, the SDE yields continuous but non-smooth trajectories $\mbf{z}_i^{0:t}$. Practically, we usually work with the Euler-Maruyama scheme for time discretization in order to solve the SDE system. Suppose we have $L+1$ time points $t^0, t^1, \cdots, t^L = T$ in the period $[0, T]$, they are equally spaced with time window $\Delta t = T/L$. Then we have a generative transition between two conservative states
\begin{align*}
\mbf{z}_i^{l+1} = \mbf{z}_i^{l} + \bm{\mu}_i^{l} \Delta t + \mathrm{diag}[(\bm{\nu}_i^{l})^{1/2}] \sqrt{\Delta t} \mathcal{N}(\mbf{0}, \mbf{I}), \, 0 \le l < L-1.
\end{align*} 
This is equivalent to the following Gaussian transition, given $\bm{\Sigma}_i^l = \mathrm{diag}[\bm{\nu}_i^l]$, as
\begin{align}
p(\mbf{z}_i^{l+1}|\mbf{z}_i^{l}) = \mathcal{N}(\mbf{z}_i^{l+1}|\mbf{z}_i^{l} + \bm{\mu}_i^{l} \Delta t, \bm{\Sigma}_i^{l} \Delta t).
\end{align}
Note that though the transition is Gaussian, the SDE finally outputs expressive posterior $q(\mbf{z}^L|\mbf{x})$ rather than the simple Gaussian, at the cost of however having no closed-form expression for $q(\mbf{z}^L|\mbf{x})$. But the related samples can be obtained by solving the SDE system via the Euler-Maruyama method.

As for the drift and diffusion, alternatively, they could be represented by the mean and variance of sparse GP to describe the SDE field~\cite{hegde2019deep}, resulting in analytical KL terms in ELBO, see Appendix~\ref{app_diffgp}. In order to enhance the representation learning, we herein build a more powerful SDE with NN-formed drift and diffusion, called neural SDE (NSDE). This however comes at the cost of intractable KL term. It is found that the SDE transformation gives the following ELBO as
\begin{align} \label{eq_elbo_sde}
\begin{split}
\mathcal{L}_{\mathrm{sde}} = &\mathbb{E}_{q(\mbf{f}|\mbf{z}^L) q(\mbf{z}^L|\mbf{x})} [\log p(\mbf{y}|\mbf{f})] \\
&- \mathrm{KL}[q(\mbf{z}^L|\mbf{x}) || p(\mbf{z}^L|\mbf{x})] - \mathrm{KL}[q(\mbf{u}) || p(\mbf{u})].
\end{split}
\end{align}
Different from the ELBO in~\eqref{eq_gpvae_elbo} which poses a Gaussian assumption for $q(\mbf{z}|\mbf{x})$, the second KL term in the right-hand side of $\mathcal{L}_{\mathrm{sde}}$ is now intractable due to the implicit density $q(\mbf{z}^L|\mbf{x})$. Alternatively, it can be estimated through the obtained $s$ SDE trajectory samples as
\begin{align}
\mathrm{KL}[q(\mbf{z}_i^L|\mbf{x}_i) || p(\mbf{z}_i^L|\mbf{x}_i)] \approx \frac{1}{s} \sum_{j=1}^s \log \frac{q(\mbf{z}_i^{L(j)}|\mbf{x}_i)}{p(\mbf{z}_i^{L(j)}|\mbf{x}_i)}.
\end{align}
To estimate the implicit density $q(\mbf{z}_i^{L}|\mbf{x}_i)$, Chen et al.~\cite{chen2018continuous} used the simple empirical method according to the SDE samples as $q(\mbf{z}_i^{L}|\mbf{x}_i) = \sum_{j=1}^s \delta(\mbf{z}_i^{L} - \mbf{z}_i^{L(j)}) / s$, where $\delta(\mbf{z}_i^{L} - \mbf{z}_i^{L(j)})$ is a point mass at $\mbf{z}_i^{L(j)}$. The estimation quality of this empirical method however is not guaranteed. It is found that since
\begin{align}
q(\mbf{z}_i^{L}|\mbf{x}_i) = \mathbb{E}_{q(\mbf{z}_i^{L-1}|\mbf{x}_i)} [q(\mbf{z}_i^{L}|\mbf{z}_i^{L-1}) ],
\end{align}
we could evaluate the density through the SDE trajectory samples from the previous time $t^{L-1}$ as
\begin{align}
\begin{split}
q(\mbf{z}_i^{L}|\mbf{x}_i) &\approx \frac{1}{s} \sum_{j=1}^{s} \mathcal{N}(\mbf{z}_i^{L}|\mbf{z}_i^{(L-1)(j)} + \bm{\mu}_i^{(L-1)(j)} \Delta t, \bm{\Sigma}_i^{(L-1)(j)} \Delta t).
\end{split}
\end{align}
In practice, we adopt the single-sample approximation together with the reparameterization trick to perform backprogate and have an unbiased estimation of the gradients~\cite{kingma2013auto}.

\subsection{Flexible prior $p(\mbf{z}|\mbf{x})$} \label{sec_flexible_prior}

\subsubsection{How about an i.i.d prior?}
Without apriori knowledge, we could simply choose an \textit{i.i.d} prior $p(\mbf{z}|\mbf{x}) = \prod_{i=1}^n \mathcal{N}(\mbf{z}_i|\mbf{0}, \mbf{I})$ with isotropic unit Gaussians. This kind of prior however is found to impede our model.

As stated before, when $\mbf{x}$ is unobservable, the bound~\eqref{eq_VAE_unsupervised_elbo} becomes the VAE-type ELBO for unsupervised learning. From the view of VAE, it is found that this ELBO may not guide the model to learn a good latent representation. The VAE is known to suffer from the issue of posterior collapse (also called KL vanishing)~\cite{chen2017variational}. That is, the learned posterior is independent of the input $\mbf{x}$, i.e., $q(\mbf{z}|\mbf{x}) \approx p(\mbf{z})$. As a result, the latent variable $\mbf{z}$ does not encode any information from $\mbf{x}$. This issue is mainly attributed to the optimization challenge of VAE~\cite{he2019lagging}. It is observed that maximizing the ELBO~\eqref{eq_VAE_unsupervised_elbo} requires minimizing $\mathrm{KL}[q(\mbf{z}) || p(\mbf{z})]$, which favors the posterior collapse in the initial training stage since it gives a zero KL value. In this case, if we are using a highly expressive decoder which is capable of modeling arbitrarily data distribution, e.g., the PixelCNN~\cite{oord2016conditional}, then the decoder will not use information from $\mbf{z}$.

As for our model, though the GP decoder is not so highly expressive, it cannot escape from the posterior collapse due to the property of GP. Furthermore, the posterior collapse of our model has been observed even in supervised learning scenario, see Fig.~\ref{fig_toy_regression}. We will prove below that the DLVKL using the simple \textit{i.i.d} prior suffers from a non-trivial state when posterior collapse happens.

Before proceeding, we make some required clarifications. First, it is known that the positions $\widetilde{\mbf{z}}$ of inducing variables $\mbf{u}_d$ fall into the latent space $\mathcal{Z} \in R^{d_{\mbf{z}}}$. They are regarded as the variational parameters of $q(\mbf{u}_d)$ and need to be optimized. However, since the latent space is not known in advance and it dynamically evolves through training, it is hard to properly initialize $\widetilde{\mbf{z}}$, which in turn may deteriorate the training quality. Hence, we employ the \textit{encoded inducing} strategy indicated as below.

\begin{definition}
	(Encoded inducing) It puts the positions $\widetilde{\mbf{x}}$ of inducing variables $\mbf{u}_d$ into the original input space $\mathcal{X}$ instead of $\mathcal{Z}$, and then passes them through the encoder to obtain the related inducing positions in the latent space as
	\begin{align*}
	\mathrm{vec}[\widetilde{\mbf{z}}] = \mathcal{N}(\mathrm{vec}[\widetilde{\mbf{z}}] | \mathtt{Encoder}(\widetilde{\mbf{x}}), \mathrm{diag}[\mathtt{Encoder}(\widetilde{\mbf{x}})]).
	\end{align*}
	Now the inducing positions $\widetilde{\mbf{z}}$ take into account the characteristics of latent space through encoder and become Gaussians.
\end{definition}

Second, the GP part employs the stationary kernel, e.g., the RBF kernel, for learning. The properties of stationary kernel are indicated as below.

\begin{definition}
	(Stationary kernel) The stationary kernel for GP is a function of the relative distance $\bm{\tau} = \mbf{x} - \mbf{x}'$. Specifically, it is expressed as $k(\bm{\tau}) = h^2 g_{\bm{\psi}}(\bm{\tau})$, where $\bm{\psi}$ is the kernel parameters which mainly control the smoothness along dimensions, and $h^2$ is the output-scale amplitude. Particularly, the stationary kernel satisfies $k(\mbf{0}) = h^2$ and $k(\bm{\tau}) = k(-\bm{\tau})$.
\end{definition}

Thereafter, the following proposition reveals that the DLVKL using \textit{i.i.d} prior fails when posterior collapse happens.

\begin{prop} \label{Prop_DLVKL}
	Given the training data $\mathcal{D} = \{\mbf{x}, \mbf{y}\}$, we build a DLVKL model using the i.i.d prior $p(\mbf{z}|\mbf{x})$, the stationary kernel and the encoded inducing strategy. When the posterior collapse happens in the initial stage of the training process, the DLVKL falls into a non-trivial state: it degenerates to a constant predictor, and the optimizer can only calibrate the prediction variance. 
\end{prop}

Detailed proof of this proposition is provided in Appendix~\ref{app_proof}. Furthermore, it is found that for any \textit{i.i.d.} prior $p(\mbf{z}_i|\mbf{x}_i) = \mathcal{N}(\bm{\mu}_0, \bm{\nu}_0)$, the degeneration in Proposition~\ref{Prop_DLVKL} happens due to the collapsed kernel matrices.  The above analysis indicates that the simple \textit{i.i.d.} prior impedes our model when posterior collapse happens. Though recently more informative priors have been proposed, for example, the mixture of Gaussians prior and the VampPrior~\cite{tomczak2018vae}, they still belong to the \textit{i.i.d.} prior. This motivates us to come up with flexible and informative prior in order to avoid the posterior collapse.

\subsubsection{A hybrid prior brings adjustable regularization}
Interestingly, we could set the prior drift $\bm{\mu} = \mbf{0}$ and diffusion $\bm{\Sigma} = \nu_0 \mbf{I}$, and let $\mbf{z}$ pass through the SDE system to have an analytical prior at time $T$ as
\begin{align} \label{eq_sde_prior}
\begin{split}
p_{\mathrm{sde}}(\mbf{z}_i^L|\mbf{x}_i) &= \mbf{x}_i + \int_0^T \mbf{0} d\tau + \int_0^T \sqrt{\nu_0} \mbf{I} d\mbf{w}^{\tau}\\ 
&= \mathcal{N}(\mbf{z}_i^L|\mbf{x}_i, \nu_0 T \mbf{I}).
\end{split}
\end{align}
The \textit{independent but not identically distributed} SDE prior is more informative than $\mathcal{N}(\mbf{0}, \mbf{I})$. For this SDE prior,
\begin{itemize}
\item it varies over data points, and will not incur the collapsed kernel matrices like the \textit{i.i.d} prior, thus sidestepping the posterior collapse, see the empirical demonstration in Fig.~\ref{fig_toy_regression}; and
\item the SDE prior performs like the skip connection~\cite{he2016deep} by connecting the original input $\mbf{x}$ to the latent output $\mbf{z}^L$ in order to avoid  rank pathologies in deep models~\cite{duvenaud2014avoiding}.
\end{itemize}
Besides, the additional variance $\nu_0$ in the SDE prior is used to build connection to the posterior. For the posterior $q(\mbf{z}_i^L|\mbf{x}_i)$, we employ the following drift and diffusion for transition $q(\mbf{z}_i^{l+1}|\mbf{z}_i^l)$ as
\begin{align} \label{eq_mlp_drift_diff}
\begin{split}
\bm{\mu}_i^l &= \mathtt{Linear}(\mathtt{MLP}(\mbf{z}_i^l)),\\
\bm{\Sigma}^l_i &= \mathrm{diag}[\nu_0 \times \mathtt{Sigmoid}(\mathtt{MLP}(\mbf{z}_i^l))].
\end{split}
\end{align}
The connection through $\nu_0$ allows knowledge transfer between the prior and the posterior, thus further improving the flexibility of the SDE prior. 

More generally, it is found that for the independent but not identically distributed prior $p(\mbf{z}_i^L|\mbf{x}_i) = \mathcal{N}(\mbf{z}_i^L|\bm{\mu}_i, \bm{\nu}_i)$, the optimal choice for maximizing ELBO is $p(\mbf{z}_i^L|\mbf{x}_i) \triangleq q(\mbf{z}_i^L|\mbf{x}_i)$. But this will cancel the KL regularizer and risk severe over-fitting. Alternatively, we could construct a composited prior, like~\cite{hoffman2017beta}, as
\begin{align} \label{eq_beta_prior}
p_{\beta}(\mbf{z}^L_{i}|\mbf{x}_{i}) = \frac{1}{r_i} q^{1-\beta}(\mbf{z}^L_{i}|\mbf{x}_{i}) p_{\mathrm{sde}}^{\beta}(\mbf{z}^L_{i}|\mbf{x}_{i}),
\end{align}
where $\beta \in [0,1]$ is a trade-off parameter, and $r_i = \int q^{1-\beta}(\mbf{z}^L_{i}|\mbf{x}_{i}) p_{\mathrm{sde}}^{\beta}(\mbf{z}^L_{i}|\mbf{x}_{i}) d\mbf{z}_i^L$ is a normalizer. When $\beta = 1$, we are using the SDE prior; when $\beta$ is a mild value, we are using a hybrid prior taking information from both the SDE prior and the variational posterior; when $\beta = 0$, we are using the variational posterior as prior.

The mixed prior gives the KL term regarding $\mbf{z}$ in $\mathcal{L}_{\mathrm{sde}}$ as
\begin{align}
\begin{split}
&\mathrm{KL}[q(\mbf{z}^L_{i}|\mbf{x}_{i}) || p_{\beta}(\mbf{z}^L_{i}|\mbf{x}_{i})] \\
=& \beta \mathrm{KL}[q(\mbf{z}^L_{i}|\mbf{x}_{i}) || p_{\mathrm{sde}}(\mbf{z}^L_{i}|\mbf{x}_{i})] - \log r_i.
\end{split}
\end{align}
Note that the term $\log r_i$ has no trainable parameters. Thereafter, the ELBO rewrites to
\begin{align} \label{eq_elbo_beta_sde}
\begin{split}
\mathcal{L}_{\mathrm{sde}}^{\beta} =& \mathbb{E}_{q(\mbf{f}|\mbf{z}^L) q(\mbf{z}^L|\mbf{x})} [\log p(\mbf{y}|\mbf{f})] \\
-& \beta \mathrm{KL}[q(\mbf{z}^L|\mbf{x}) || p_{\mathrm{sde}}(\mbf{z}^L|\mbf{x})] - \mathrm{KL}[q(\mbf{u}) || p(\mbf{u})].
\end{split}
\end{align}
The formulation of $\beta$-ELBO in~\eqref{eq_elbo_beta_sde} on the other hand indicates that $\beta$ can be interpreted as a trade-off between the likelihood (data fit) and the KL regularizer w.r.t $\mbf{z}^L$, which is similar to~\cite{higgins2017beta}. This raises an adjustable regularization on the latent representation $\mbf{z}^L$: when $\beta = 1$, the SDE prior poses the most strict regularization; when $\beta = 0$, the optimal prior ignores the KL regularizer, like DKL, and focuses on fitting the training data. It is recommended to use an intermediate $\beta$, e.g., $10^{-2}$, to achieve a good trade-off; or an annealing-$\beta$ strategy~\cite{sonderby2016train, fu2019cyclical}. Note that the $\beta$-ELBO can also be derived from the view of variational information bottleneck (VIB)~\cite{alemi2016deep}, see Appendix~\ref{app_vib}.

\subsection{Discussions}
Fig.~\ref{fig_DLVKL} illustrates the structure of the proposed DLVKL and DLVKL-NSDE. It is found that the DLVKL is a special case of DLVKL-NSDE: when $T = 1.0$ and $L = 1$, $q(\mbf{z}^{L}|\mbf{x})$ becomes Gaussian. Compared to the DLVKL, the DLVKL-NSDE generates more expressive variational posterior through NSDE, which is beneficial to further reduce the gap to the exact posterior. However, the NSDE transformation requires that the states $\{\mbf{z}^l \}_{l=1}^L$ should have the same dimensionality over time. It makes DLVKL-NSDE unsuitable for handling high-dimensional data, e.g., images. This however is not an issue for DLVKL since it has no limits on the dimensionality of $\mbf{z}$. Besides, it is notable that when the encoder has $d_{\mbf{z}} < d_{\mbf{x}}$, we cannot directly use the SDE prior~\eqref{eq_sde_prior}. Alternatively, we could apply some simple dimensionality reduction algorithms, e.g., the principle component analysis (PCA), on the inputs to obtain the prior mean. When $d_{\mbf{z}} > d_{\mbf{x}}$, we use the zero-padding strategy to obtain the prior mean. In the experiments below, unless otherwise indicated, we use the DLVKL-NSDE with $d_{\mbf{x}} = d_{\mbf{z}}$, $\beta = 10^{-2}$, $T=1.0$ and $L=10$. When predicting, we average over $s=10$ posterior samples.

\begin{figure}[t!]
	\centering
	\includegraphics[width=0.5\textwidth]{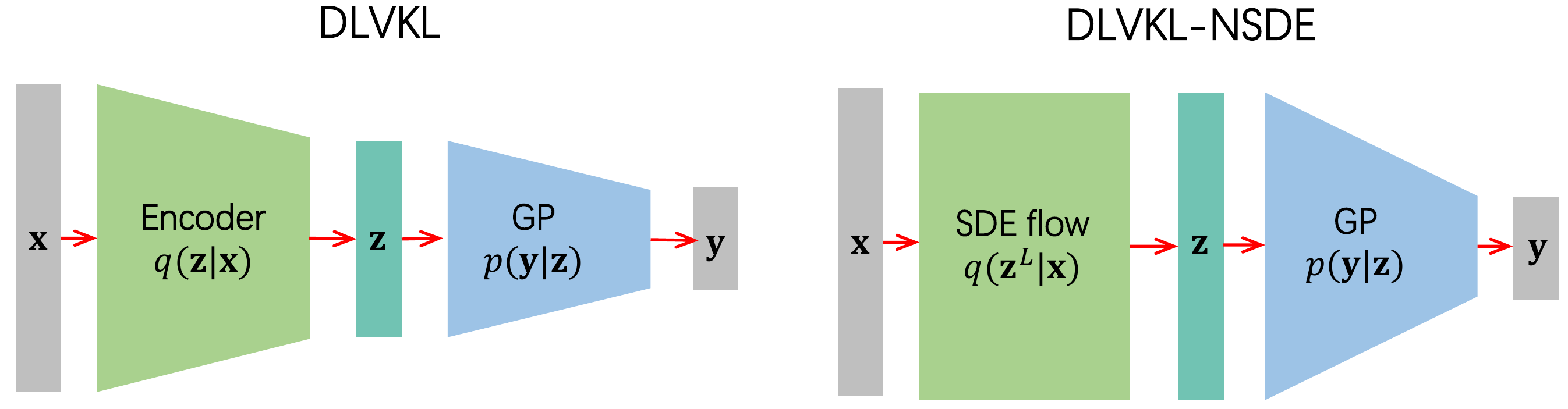}
	\caption{The model structure of the proposed DLVKL and DLVKL-NSDE.}
	\label{fig_DLVKL}
\end{figure}

\section{Numerical experiments} \label{sec_exp}
This section first uses two toy cases to investigate the characteristics of the proposed models, followed by intensive evaluation on nine regression and classification datasets. Finally, we also simply showcase the capability of unsupervised DLVKL on the $\mathtt{mnist}$ dataset. The model configurations in the experimental study are detailed in Appendix~\ref{app_exp_details}. All the experiments are performed on a windows workstation with twelve 3.50 GHz core and 64 GB memory.

\subsection{Toy cases} \label{sec_toy}
This section seeks to investigate the characteristics of the proposed models on two toy cases. Firstly, we illustrate the benefits brought by the flexible prior $p_{\beta}(\mbf{z}^L|\mbf{x})$ and the expressive posterior $q(\mbf{z}^L|\mbf{x})$ on a regression case expressed as
\begin{align*}
y(x) = \left \{\begin{array}{rr}
\cos(5x) \times \exp(-0.5x)+1, & x < 0, \\
\cos(5x) \times \exp(-0.5x)-1, & x \ge 0,
\end{array} 
\right.
\end{align*}
which has a step behavior at the origin. The conventional GP using stationary kernels is hard to capture this non-stationary behavior. We draw 50 points from $[-1.5, 1.5]$ together with their observations as training data. 

\begin{figure*}[t!]
	\centering
	\includegraphics[width=0.8\textwidth]{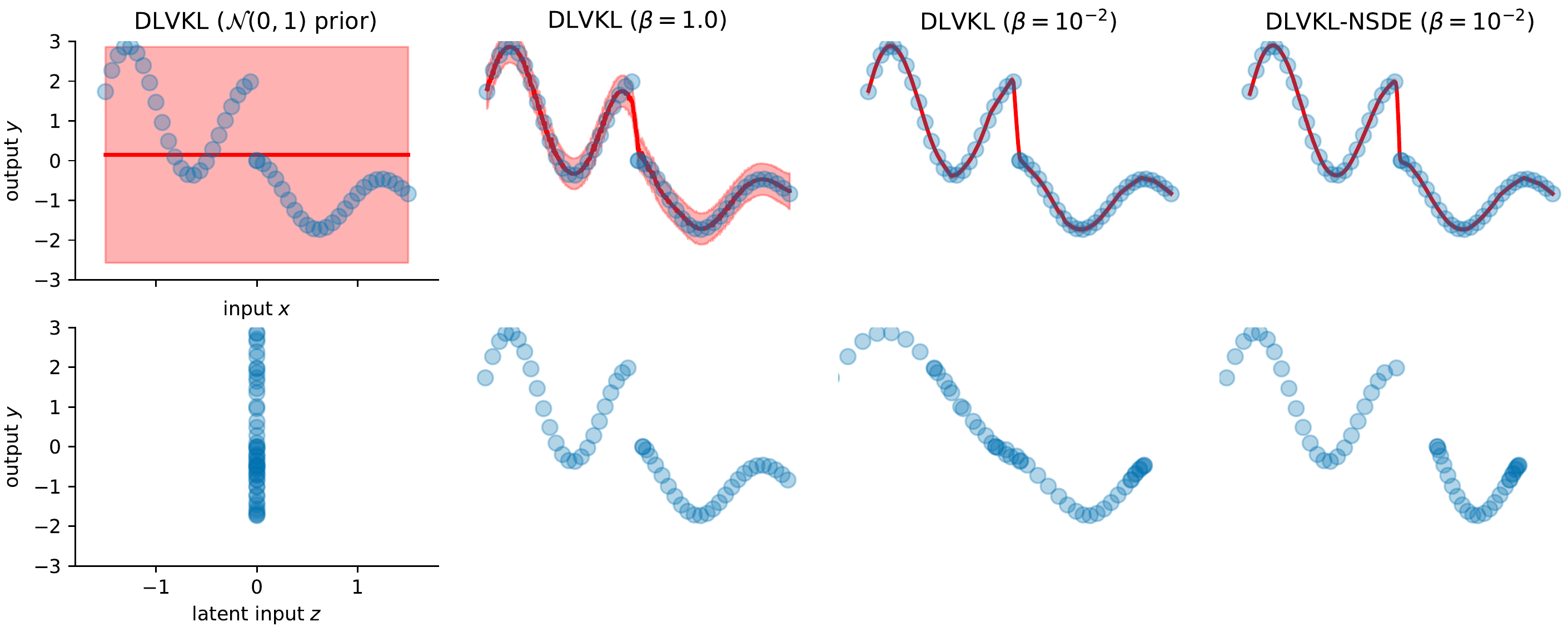}
	\caption{The variants of the proposed model under various priors and variational posteriors on the toy regression case. The blue circles in the top row represent the pairs $(\mbf{x}, \mbf{y})$ of training data, while the bottom ones are the pairs $(\mbf{z}, \mbf{y})$. The red curve is the prediction mean and the red shallow region indicates 95\% confidence interval of the prediction.}
	\label{fig_toy_regression}
\end{figure*}

Fig.~\ref{fig_toy_regression} illustrates the predictions of DLVKL and DLVKL-NSDE, together with the mean of the learned latent input $\mbf{z}$. It is found, from left to right, that the \textit{i.i.d} $\mathcal{N}(0,1)$ prior leads to collapsed posterior and constant predictor, which agree with the analysis in Proposition~\ref{Prop_DLVKL}. Instead, the independent but not identically distributed SDE prior in~\eqref{eq_sde_prior} helps the DLVKL sidestep the posterior collapse. But since this prior takes no knowledge from the posterior $q(\mbf{z}|\mbf{x})$ under $\beta = 1.0$, the DLVKL leaves almost no space for the encoder $p(\mbf{z}|\mbf{x})$ to perform deep representation learning, which is crucial for capturing the step feature. As a result, the pure SDE prior makes DLVKL perform like a GP. Hence, to improve the capability of encoder, we employ the $\beta$-mixed flexible prior in~\eqref{eq_beta_prior}, which takes both the SDE prior and the variational posterior into consideration. Now we can observe that the encoder under $\beta = 10^{-2}$ skews the original inputs in the latent space in order to make the GP part of DLVKL describe the step behavior easily. Moreover, the DLVKL-NSDE further employs the NSDE-transformed variational posterior rather than the Gaussian, thus resulting in better latent representation.

\begin{figure*}[t!]
	\centering
	\includegraphics[width=0.8\textwidth]{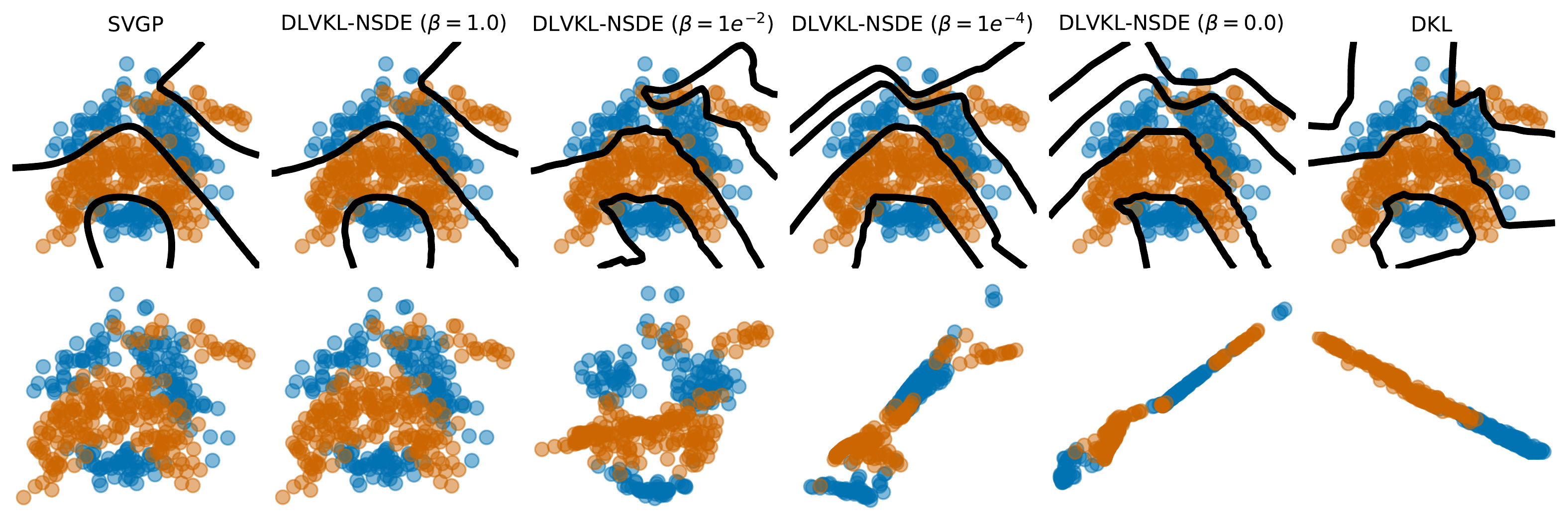}
	\caption{Varying $\beta$ changes the behavior of DLVKL-NSDE from SVGP to DKL on the toy classification case. The bottom row shows the data points transformed in the latent space $\mathcal{Z}$.}
	\label{fig_toy_classification}
\end{figure*}

Next, Fig.~\ref{fig_toy_classification} investigates the impact of $\beta$ on the behavior of DLVKL-NSDE on a toy binary classification case by changing it from $1.0$ to $10^{-4}$. We also show the results of stochastic variational GP (SVGP)~\cite{hensman2013gaussian} and DKL for comparison. It is observed that the decreasing $\beta$ changes the behavior of DLVKL-NSDE from SVGP to DKL. The decreasing $\beta$ indicates that (a) the prior contains more information from the posterior, and it has $p(\mbf{z}|\mbf{x}) = q(\mbf{z}|\mbf{x})$ when $\beta = 0$, like DKL; (b) the KL penalty w.r.t $\mbf{z}^L$ is weakened in~\eqref{eq_elbo_beta_sde}, thus risking over-fitting; and meanwhile, (c) the encoder becomes more flexible, e.g., the extreme $\beta=0$ skews the 2D inputs to an almost 1D manifold, raising the issue of rank pathologies in deep models~\cite{duvenaud2014avoiding}. In practice, we need to trade off the KL regularization for guarding against over-fitting and the representation learning for improving model capability through the $\beta$ value.

\begin{figure*}[t!]
	\centering
	\includegraphics[width=0.8\textwidth]{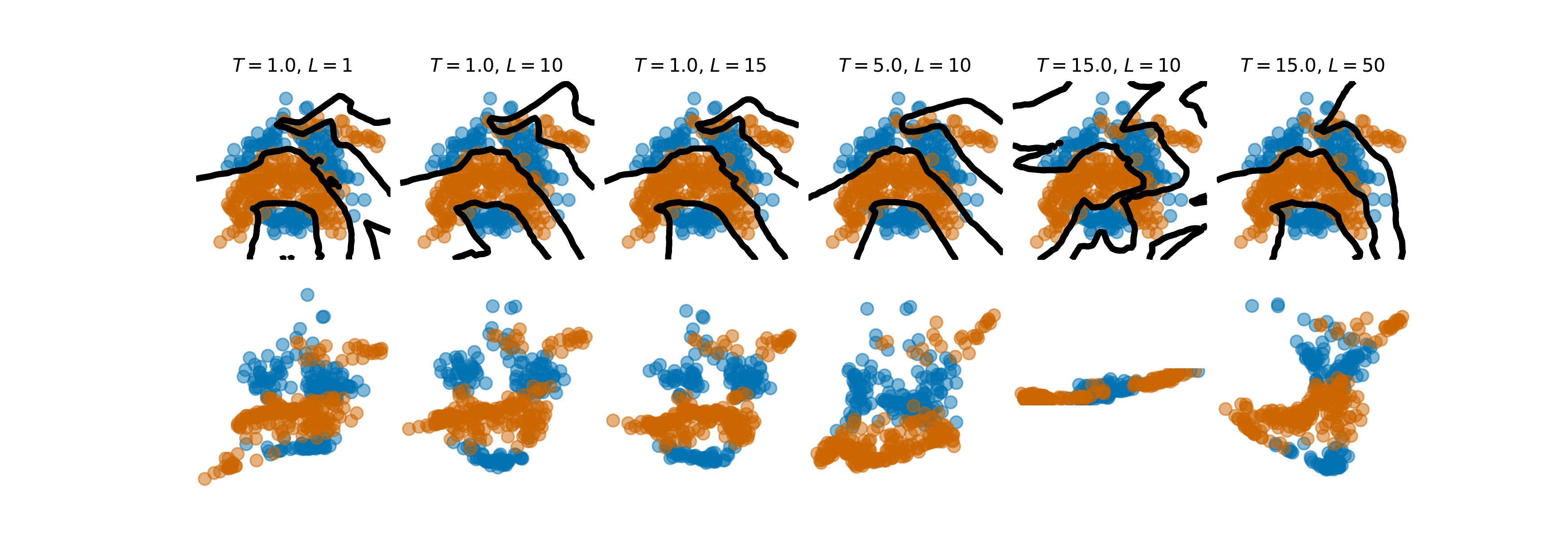}
	\caption{Impact of the flow time $T$ and flow step $L$ on the performance of DLVKL-NSDE on the toy classification case. The bottom row shows the data points transformed in the latent space $\mathcal{Z}$.}
	\label{fig_toy_flow_time_steps}
\end{figure*}

Finally, Fig.~\ref{fig_toy_flow_time_steps} investigates the impact of SDE flow parameters $T$ and $L$ on the performance of DLVKL-NSDE. We first fix the flow time as $T = 1.0$ and increase the flow step from $L = 1$ to $L = 15$. When we directly use $T=1.0$, $L = 1$ (DLVKL-NSDE herein degenerates to DLVKL), it leads to a single transition density with large variance, thus resulting in high degree of feature skewness. This in turn raises slight over-fitting in Fig.~\ref{fig_toy_flow_time_steps} as it identifies several orange points within the blue group. In contrast, the refinement of flow step makes the time discretization close to the underlying process and stabilizes the SDE solver. Secondly, we fix the flow step as $L = 10$ and increase the flow time from $T=1.0$ to $T=15.0$. This increases the time window $\Delta t$ and makes the transition density having larger variance. Hence, the encoder is equipped with higher perturbation. But purely increasing $T$ will deteriorate the quality of SDE solver, which is indicated by the issue of rank pathologies for DLVKL-NSDE with $T=15.0$, $L=10$. To summarize, in order to gain benefits from the SDE representation learning, the flow step $L$ should increase with flow time $T$. For example, in comparison to the case of $T=15.0$ and $L=10$, the DLVKL-NSDE using $T=15.0$ and $L=50$ in Fig.~\ref{fig_toy_flow_time_steps} yields reasonable predictions and latent representation.

\subsection{Regression and classification} \label{sec_exp_supervised}
We here evaluate the proposed model on six UCI regression datasets, two classification datasets and the $\mathtt{cifar}$-$\mathtt{10}$ image classification dataset summarized in Table~\ref{tab_datasets}. The data size ranges from 506 to 11M in order to conduct intensive comparison at different levels. It is notable that the first three small regression datasets in Table~\ref{tab_datasets} could help verify whether the proposed model can achieve reasonable regularization to guard against over-fitting or not.

The competitors include the pure SVGP~\cite{hensman2013gaussian} and NN, the DiffGP~\cite{hegde2019deep}, a SDE-based deep GP, and finally the DKL~\cite{wilson2016deep}. For the regression tasks, we employ the root mean square error (RMSE) and the negative log likelihood (NLL) as performance criteria to quantify the quality of predictive distribution. Similarly, we verify the performance of classification in terms of classification accuracy and NLL. Tables~\ref{tab_reg} and~\ref{tab_clasfy} report the comparative results in terms of RMSE (accuracy) and NLL, respectively.\footnote{We only provide the RMSE and accuracy results for the deterministic NN. Besides, the SVGP and DiffGP are not applied on the $\mathtt{cifar}$-$\mathtt{10}$ dataset, since the pure GPs cannot effectively handle the high-dimensional image data.}  The best and second-best results are marked in gray and light gray, respectively. Based on the comparative results, we have the following findings.

\begin{table}
	\caption{The regression and classification datasets} 
	\label{tab_datasets}
	\centering
		\begin{tabular}{lrrrr}
			\hline
			dataset &$n_{\mathrm{train}}$ &$n_{\mathrm{test}}$ &$d_{\mbf{x}}$ &no. of classes\\
			\hline
			\hline
			boston	&456 &50 &13	&-\\
			concrete &927 &103 &8 &- \\
			wine-red	&1440 &159 &22	&-\\
			keggdirected	&43945 &4882 &20	&-\\
			kin40k	&36000 &4000 &8	&-\\ 	
			protein	&41157 &4573 &9	&-\\	
			\hline
			\hline
			connect-4	&60802 &6755 &43	&2	\\
			higgs	&9900000 &1100000 &28	&2\\	
			cifar-10	&50000 &10000 &32$\times$32$\times$3	&10\\			 
			\hline
		\end{tabular}
\end{table}

\begin{figure}[t!]
	\centering
	\includegraphics[width=0.5\textwidth]{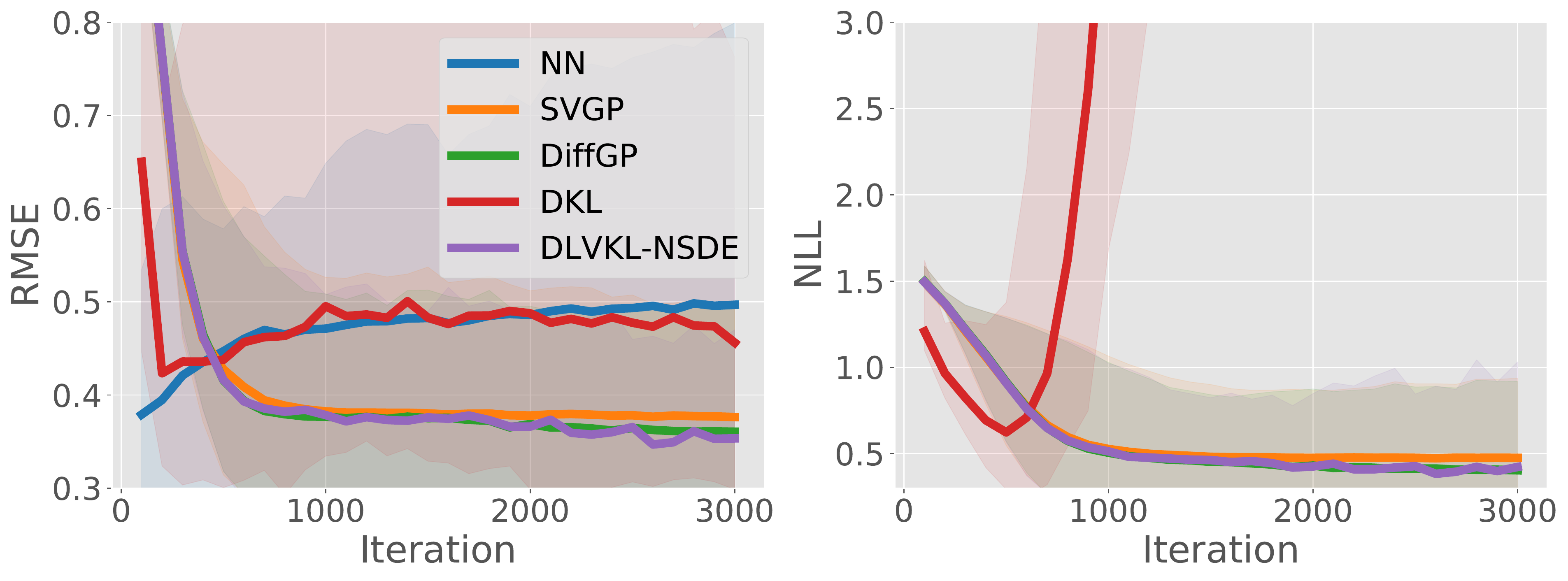}
	\caption{Comparative results on the small $\mathtt{boston}$ dataset. The curves represent the average results over ten runs, while the shallow regions are the bounds (minimum and maximum) around the mean.}
	\label{fig_boston}
\end{figure}

\textbf{The DKL risks over-fitting on small datasets.} It is not surprising that the powerful NN without regularization is over-fitted on the first three small datasets. But we observe that though equipped with a GP layer, the deterministic representation learning also risks over-fitting for DKL on the small datasets. Fig.~\ref{fig_boston} depicts the comparative results on the small $\mathtt{boston}$ dataset. It indicates that the DKL improves the prediction quickly without regularization for the deterministic encoder. But the free latent representation weakens the regularization of the following GP part. Consequently, over-fitting occurs after around 200 iterations and severely underestimated prediction variance happens after 500 iterations. 

Besides, the DKL directly optimizes the positions of inducing points in the dynamic, unknown latent space $\mathcal{Z}$. Without prior knowledge about $\mathcal{Z}$, we can only use the inputs $\mbf{x}$ to initialize $\widetilde{\mbf{z}}$. The mismatch of data distributions in the two spaces $\mathcal{X}$ and $\mathcal{Z}$ may deteriorate and even lead to inappropriate termination on the training process of DKL. For instance, the DKL fails in several runs on the $\mathtt{kin40k}$ dataset, indicated by the high standard deviations in the tables.

\textbf{The DLVKL-NSDE achieves a good trade-off.} Different from the DKL, the proposed DLVKL-NSDE builds the whole model in the Bayesian framework. Due to the scarce training data, we use $\beta = 1.0$ for DLVKL-NSDE on the first three datasets in order to completely use the KL regularizer w.r.t $\mbf{z}^L$ in~\eqref{eq_elbo_beta_sde}. As a result, the DLVKL-NSDE performs similarly to the well-calibrated GP and DiffGP on the first three small datasets, see Tables~\ref{tab_reg} and~\ref{tab_clasfy} and Fig.~\ref{fig_boston}. As for the six large datasets, it is not easy to incur over-fitting. Hence, the DLVKL-NSDE employs a small balance parameter $\beta=10^{-2}$ to fully use the power of deep latent representation, and consequently it shows superiority on four datasets. 

\textbf{Deep latent representation improves the prediction on large datasets.} It is observed that in comparison to the pure SVGP, the deep representation learning improves the capability of DiffGP, DKL and DLVKL-NSDE, thus resulting in better performance on most cases. Besides, in comparison to the sparse GP assisted representation learning in DiffGP, the more flexible NN based representation learning further enhances the performance of DKL and DLVKL-NSDE, especially on large datasets.

\begin{table*}
	\caption{The RMSE results for regression and the accuracy results for classification. For the RMSE criterion, lower is better; while for the accuracy criterion, higher is better.} 
	\label{tab_reg}
	\centering
		\begin{tabular}{lrrrrr}
			\hline
			dataset &NN &SVGP & DiffGP & DKL & DLVKL-NSDE \\
			\hline
			boston	&0.4953$_{\pm0.1350}$	&0.3766$_{\pm0.0696}$	&\cellcolor{mygray}0.3629$_{\pm0.0668}$
			&0.4703$_{\pm0.1748}$	&\cellcolor{lightgray}0.3476$_{\pm0.0745}$	\\
			concrete	&0.3383$_{\pm0.0314}$	&0.3564$_{\pm0.0250}$	&\cellcolor{lightgray}0.3232$_{\pm0.0303}$
            &0.3520$_{\pm0.0471}$	&\cellcolor{mygray}0.3375$_{\pm0.0278}$	\\
			wine-red &0.9710$_{\pm0.0748}$	&0.7779$_{\pm0.0382}$	&\cellcolor{mygray}0.7779$_{\pm0.0381}$
            &0.9414$_{\pm0.0974}$	&\cellcolor{lightgray}0.7779$_{\pm0.0380}$	\\
			keggdirected	&0.1125$_{\pm0.0820}$	&0.0924$_{\pm0.0053}$	&0.0900$_{\pm0.0054}$	&\cellcolor{mygray}0.0894$_{\pm0.0052}$	&\cellcolor{lightgray}0.0875$_{\pm0.0057}$	\\
			kin40k	&\cellcolor{mygray}0.1746$_{\pm0.0109}$	&0.2772$_{\pm0.0043}$	&0.2142$_{\pm0.0044}$	&0.7519$_{\pm0.3751}$	&\cellcolor{lightgray}0.1054$_{\pm0.0048}$	\\ 	
			protein	&\cellcolor{mygray}0.6307$_{\pm0.0069}$	&0.7101$_{\pm0.0090}$	&0.6763$_{\pm0.0104}$	&0.6452$_{\pm0.0084}$	&\cellcolor{lightgray}0.6098$_{\pm0.0088}$	\\	
			\hline
			connect-4	&0.8727$_{\pm0.0037}$	&0.8327$_{\pm0.0030}$	&0.8550$_{\pm0.0045}$	&\cellcolor{mygray}0.8756$_{\pm0.0017}$	&\cellcolor{lightgray}0.8826$_{\pm0.0032}$	\\
			higgs	&\cellcolor{lightgray}0.7616$_{\pm0.0014}$	&0.7280$_{\pm0.0004}$	&0.7297$_{\pm0.0008}$	&\cellcolor{mygray}0.7562$_{\pm0.0017}$	&0.7529$_{\pm0.0017}$	\\	
			cifar-10	&0.9155	&NA	&NA	&\cellcolor{lightgray}0.9186	&\cellcolor{mygray}0.9176	\\			 
			\hline
		\end{tabular}
\end{table*}

\begin{table*}
	\caption{The NLL results for regression and classification. For this criterion, lower is better.} 
	\label{tab_clasfy}
	\centering
	\centering
		\begin{tabular}{lrrrr}
			\hline
			dataset &SVGP & DiffGP & DKL & DLVKL-NSDE \\
			\hline
			boston	&0.4752$_{\pm0.2377}$	&\cellcolor{mygray}0.4081$_{\pm0.2299}$	&48.6696$_{\pm57.1183}$	&\cellcolor{lightgray}0.3792$_{\pm0.2550}$	\\
			concrete	&0.3759$_{\pm0.0657}$	&\cellcolor{lightgray}0.2736$_{\pm0.0941}$	&1.7932$_{\pm1.1077}$	&\cellcolor{mygray}0.3295$_{\pm0.0934}$	\\
			wine-red	&\cellcolor{mygray}1.1666$_{\pm0.0475}$	&\cellcolor{lightgray}1.1666$_{\pm0.0473}$		&4.1887$_{\pm1.1973}$ &1.1668$_{\pm0.0471}$	\\
			keggdirected	&-0.9975$_{\pm0.0321}$	&\cellcolor{mygray}-1.0283$_{\pm0.0357}$	&-1.0245$_{\pm0.0360}$	&\cellcolor{lightgray}-1.0568$_{\pm0.0349}$	\\
			kin40k	&0.1718$_{\pm0.0092}$	&\cellcolor{mygray}-0.0853$_{\pm0.0125}$	&0.9002$_{\pm0.7889}$	&\cellcolor{lightgray}-0.8456$_{\pm0.0483}$	\\ 	
			protein	&1.0807$_{\pm0.0116}$	&1.0298$_{\pm0.0142}$	&\cellcolor{mygray}0.9817$_{\pm0.0132}$	&\cellcolor{lightgray}0.9258$_{\pm0.0155}$	\\	
			\hline
			connect-4	&0.3637$_{\pm0.0049}$	&0.3207$_{\pm0.0058}$	&\cellcolor{mygray}0.3098$_{\pm0.0101}$	&\cellcolor{lightgray}0.2812$_{\pm0.0076}$	\\
			higgs	&0.5356$_{\pm0.0004}$	&0.5325$_{\pm0.0016}$	&\cellcolor{lightgray}0.4907$_{\pm0.0021}$	&\cellcolor{mygray}0.4980$_{\pm0.0025}$	\\	
			cifar-10	&NA	&NA	&\cellcolor{lightgray}0.3546	&\cellcolor{mygray}0.3710	\\			 
			\hline
		\end{tabular}
\end{table*}

\begin{figure}[t!]
	\centering
	\includegraphics[width=0.5\textwidth]{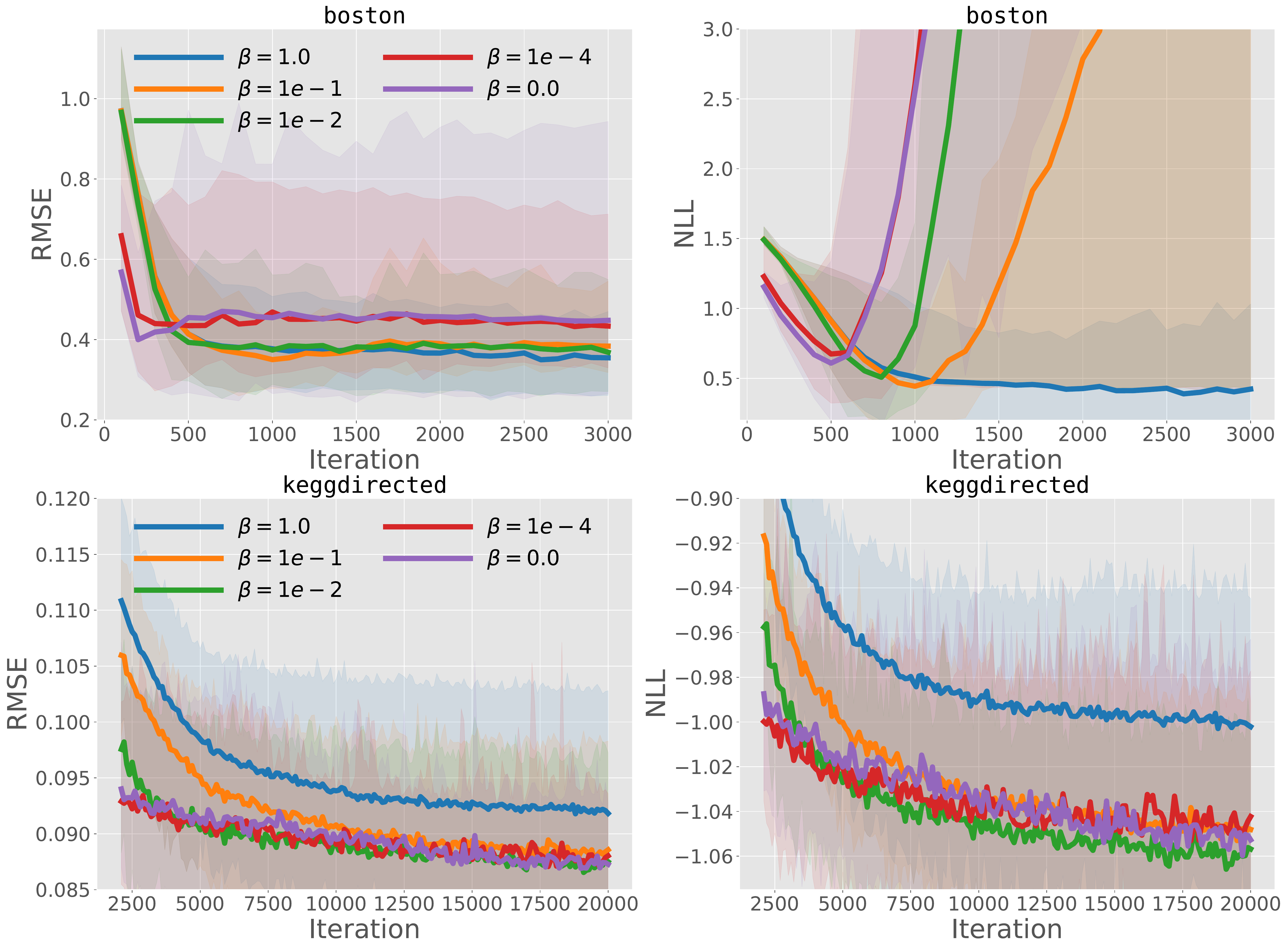}
	\caption{Impact of the parameter $\beta$ on the performance of DLVKL-NSDE on the small $\mathtt{boston}$ dataset and the large $\mathtt{keggdirected}$ dataset.}
	\label{fig_beta_boston_keggdirected}
\end{figure}

\textbf{Impact of $\beta$ and flow parameters.} Finally, we discuss the impact of $\beta$ and the SDE flow parameters $T$ and $L$ on the performance of DLVKL-NSDE. As for the trade-off parameter $\beta$, which adjusts the flexibility of prior $p_{\beta}(\mbf{z}^L|\mbf{x})$ and the weight of the KL regularizer $\mathrm{KL}[q(\mbf{z}^L|\mbf{x}) || p_{\mathrm{sde}}(\mbf{z}^L|\mbf{x})]$, Fig.~\ref{fig_beta_boston_keggdirected} performs investigation on the small $\mathtt{boston}$ and the medium-scale $\mathtt{keggdirected}$ datasets by varying $\beta$ from 1.0 to 0.0. The decrease of $\beta$ improves the flexibility of the hybrid prior since it takes into account more information from the posterior $q(\mbf{z}^L|\mbf{x})$ through~\eqref{eq_beta_prior}. Meanwhile, it weakens the role of KL regularizer to improve the freedom of representation learning, which is beneficial to minimize the first likelihood term of~\eqref{eq_elbo_beta_sde}. Consequently, small $\beta$ speeds up the training of DLVKL-NSDE. But the deteriorated KL regularizer with decreasing $\beta$ makes DLVKL-NSDE be approaching the DKL. Hence, we observe over-fitting and underestimated prediction variance for DLVKL-NSDE with small $\beta$ on the $\mathtt{boston}$ dataset. As for the medium-scale $\mathtt{keggdirected}$ dataset with many more data points, the issues have not yet happened. But it is found that (i) $\beta=1.0$ is over-regularized on this dataset; and (ii) the extreme $\beta=0.0$ slights deteriorates the RMSE and NLL results. Hence, we can conclude that (i) a large $\beta$ is favored to fully regularize the model on small datasets in order to guard against over-fitting; while (ii) a small $\beta$ is recommended to improve representation learning on complicated large datasets; also it is notable that
(iii) when we are using the NN encoder with higher depth and more units, which increase the model complexity, the $\beta$ should be accordingly increased.

\begin{figure}[t!]
	\centering
	\includegraphics[width=0.5\textwidth]{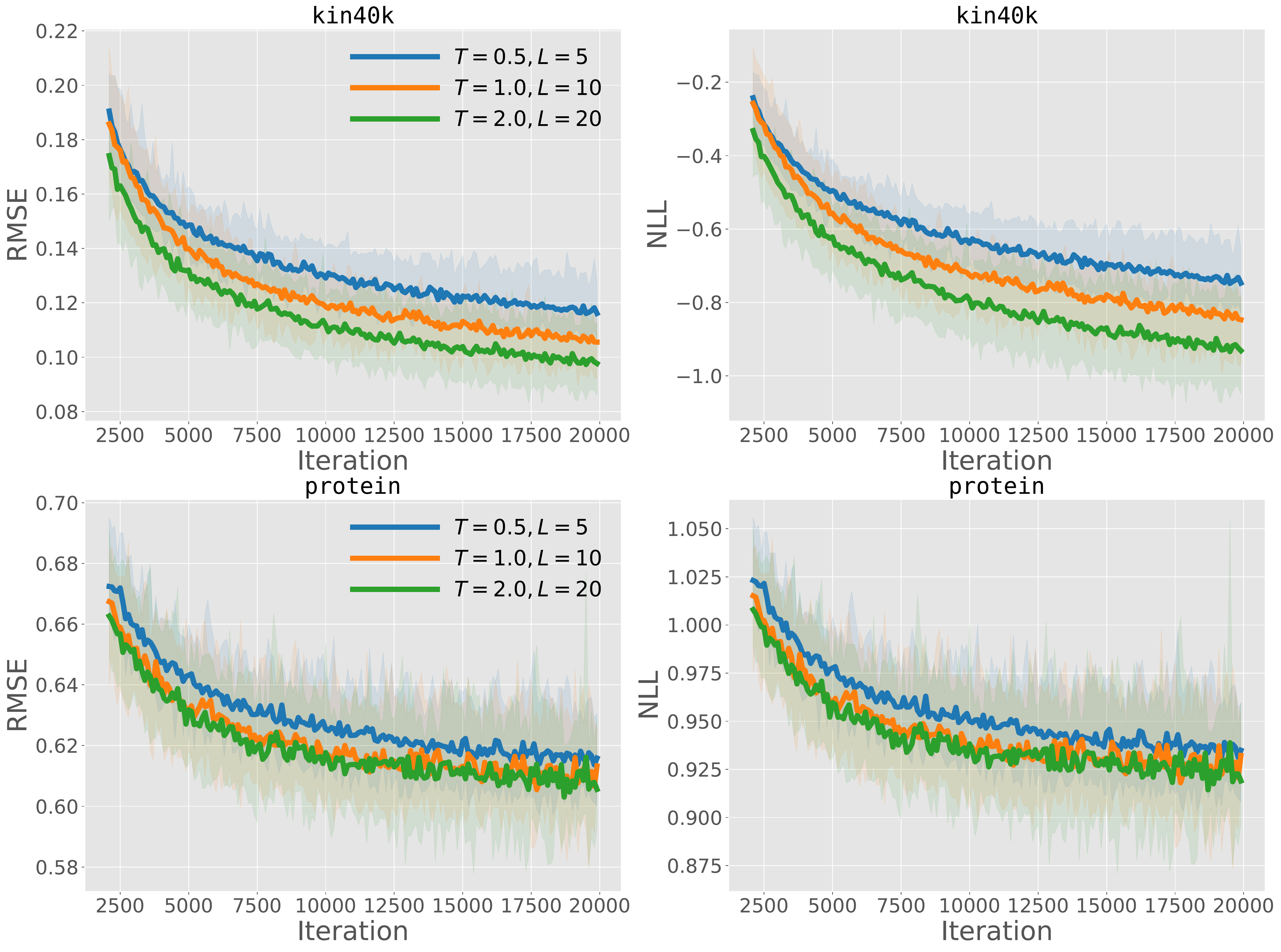}
	\caption{Impact of the SDE flow parameters $T$ and $L$ on the performance of DLVKL-NSDE on the $\mathtt{kin40k}$ and $\mathtt{protein}$ datasets.}
	\label{fig_flow_time_kin40k_protein}
\end{figure}

Fig.~\ref{fig_flow_time_kin40k_protein} studies the impact of SDE flow parameters on the $\mathtt{kin40k}$ and $\mathtt{protein}$ datasets by varying the flow time $T$ from 0.5 to 2.0. Note that according to the discussions in Section~\ref{sec_toy}, the flow step $L$ is accordingly increased  to ensure the quality of SDE solver. The longer SDE flow time transforms the inputs to a more expressive posterior $q(\mbf{z}^L|\mbf{x})$ in order to reduce the gap to the exact posterior. As a result, the DLVKL-NSDE improves the performance with increasing $T$ on the $\mathtt{kin40k}$ dataset. As for the $\mathtt{protein}$ dataset, it is found that $T = 1.0$ is enough since the longer flow time does not further improve the results. Note that the time complexity of DLVKL-NSDE increases with $T$ and $L$. As a trade-off, we usually employ $T = 1.0$ and $L = 10$ in our experiments.

\subsection{Unsupervised learning on the $\mathtt{mnist}$ dataset}
It is notable that the proposed DLVKL-NSDE can also be used for unsupervised learning once we replace the input $\mbf{x}$ with the output $\mbf{y}$ in~\eqref{eq_elbo_beta_sde}. To verify this, we apply the model to the $\mathtt{mnist}$ handwritten digit dataset, which contains 60000 gray images with size $28 \times 28$. Since the VAE-type unsupervised learning structure requires feature transformations with varying dimensions, we employ the DLVKL-NSDE using $T = 1.0$ and $L = 1$, i.e., the DLVKL. In this case, the DLVKL is similar to the GPLVM using back constraints (recognition model)~\cite{bui2015stochastic}. The difference is that DLVKL uses $\beta = 10^{-2}$ instead of $\beta = 1.0$ in GPLVM. Besides, the competitors include VAE~\cite{kingma2013auto} and DKL. The details of experimental configurations are provided in Appendix~\ref{app_exp_details}.

\begin{figure}[t!]
	\centering
	\includegraphics[width=.5\textwidth]{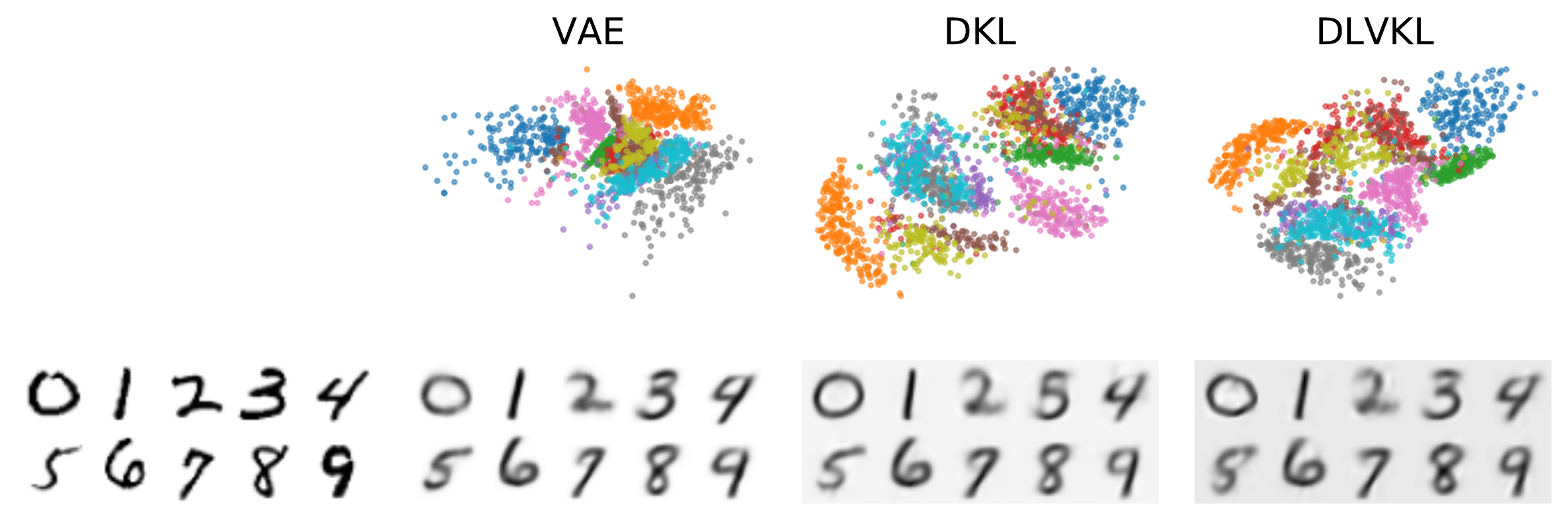}
	\caption{Unsupervised learning on the $\mathtt{mnist}$ dataset. The top row represents the learned two-dimensional latent space, while the bottom row illustrates the reconstructed digits in comparison to the ground truth (bottom left).}
	\label{fig_mnist}
\end{figure}

Fig.~\ref{fig_mnist} illustrates the two-dimensional latent space learned by different models, and several reconstructed digits. It is clearly observed that the models properly cluster the digits in the latent space. As for reconstruction, the three models reconstruct the profile of digits but lost some details due to the limited latent dimensionality. Besides, the reconstructed digits of DKL and DLVKL have slightly noisy background due to the Bayesian averaging and the share of kernel across 784 outputs. Finally, the DLVKL is found to have more reasonable reconstruction than DKL for some digits, e.g., digit ``3''.

Finally, note that for the $\mathtt{mnist}$ dataset together with the $\mathtt{cifar}$-$\mathtt{10}$ dataset in Section~\ref{sec_exp_supervised}, we use the original GP in our models. Some recently developed GP paradigms, for example, the convolutional GP~\cite{van2017convolutional} for images, could be considered to further improve the performance.

\section{Conclusions} \label{sec_con}
This paper proposes the DLVKL which inherits the advantages of DKL but provides better calibration through regularized latent representation. We further improve the DLVKL through (i) the NSDE transformation to yield expressive variational posterior, and (ii) the hybrid prior to achieve adjustable regularization on latent representation. We investigate the algorithmic characteristics of DLVKL-NSDE and compare it against existing deep GPs. The comparative results imply that the DLVKL-NSDE performs similarly to the well calibrated GP on small datasets, and shows superiority on large datasets.

\section*{Acknowledgments}
This work was supported by the Fundamental Research Funds for the Central Universities (DUT19RC(3)070) at Dalian University of Technology, and it was partially supported by the Research and Innovation in Science and Technology Major Project of Liaoning Province (2019JH1-10100024), the National Key Research and Development Project (2016YFB0600104), and the MIIT Marine Welfare Project (Z135060009002).

\appendices
\section{Viewing existing deep GPs in the framework of~\eqref{eq_dlvkl_model}} \label{app_dgps}
\subsection{Deterministic representation learning of $\mbf{z}$} \label{app_dkl}
The DKL~\cite{wilson2016deep} has the transformation from $\mbf{x}_i$ to $\mbf{z}_i$ performed through a deterministic manner $\mbf{z}_i = \mathtt{MLP}(\mbf{x}_i)$.
As a result, the ELBO is expressed as
\begin{align*}
\mathcal{L}_{\mathrm{dkl}} = \mathbb{E}_{q(\mbf{f}|\mbf{z}=\mathtt{MLP}(\mbf{x}))} [\log p(\mbf{y}|\mbf{f})] - \mathrm{KL}(q(\mbf{u}) || p(\mbf{u})).
\end{align*}
Different from~\eqref{eq_gpvae_elbo}, the purely deterministic transformation in the above ELBO will risk over-fitting, especially on small datasets, which has been verified in our numerical experiments.

\subsection{Bayesian representation learning of $\mbf{z}$ via GPs}
Inspired by NN, the DGP~\cite{damianou2013deep} extends $p(\mbf{z}|\mbf{x})$ to a $L$-layer hierarchical structure, wherein each layer is a sparse GP, as
\begin{align*}
p(\mbf{z}^{1:L}|\mbf{x}) = \prod_{l=1}^{L} p(\mbf{z}^{l}|\mbf{z}^{l-1}) = \prod_{l=1}^{L} p(\mbf{z}^{l}|\mbf{u}^l, \mbf{z}^{l-1}) p(\mbf{u}^l),
\end{align*}
where $\mbf{z}^0 = \mbf{x}$. As a result, the ELBO is expressed as
\begin{align*}
\begin{split}
\mathcal{L}_{\mathrm{dgp}} =& \mathbb{E}_{q(\mbf{f}|\mbf{z}^L) q(\mbf{z}^L|\mbf{x})} [\log p(\mbf{y}|\mbf{f})] \\
&- \sum_{l=1}^L \mathrm{KL}[q(\mbf{u}^l) || p(\mbf{u}^l)] - \mathrm{KL}[q(\mbf{u}) || p(\mbf{u})],
\end{split}
\end{align*}
where $q(\mbf{z}^L|\mbf{x}) = \int \prod_{l=1}^L p(\mbf{z}^{l}|\mbf{u}^l, \mbf{z}^{l-1}) q(\mbf{u}^l) d\mbf{u}^{1:L} d\mbf{z}^{1:L-1}  = \int \prod_{l=1}^L q(\mbf{z}^{l}|\mbf{z}^{l-1}) d\mbf{z}^{1:L-1}$. Due to the complete GP paradigm, the DGP naturally guards against over-fitting. But the capability of representation learning is limited by the finite inducing variables $\{\mbf{u}^l \}_{l=1}^L$ for massive data.

\subsection{Bayesian representation learning of $\mbf{z}$ via SDE} \label{app_diffgp}
Different from traditional DGP, the sequence of transformation of which is indexed on discrete domain, the differential GP (DiffGP)~\cite{hegde2019deep} generalizes the transformation through the SDE indexed on continuous-time domain. Specifically, given the same dimension $d^l = d^{l-1} = d_{\mbf{z}}$, $1 \le l \le L$, the posterior transformation $q(\mbf{z}_i^{l}|\mbf{z}_i^{l-1})$ through a sparse GP can be extended and interpreted as a SDE as
\begin{align*}
d\mbf{z}_i^t = \bm{\mu}^t_i + \bm{L}^t_i d\mbf{w}^t, \quad t \in [0, T],
\end{align*}
where $\bm{\Sigma}^t_i = \bm{L}^t_i (\bm{L}^t_i)^{\mathsf{T}}$ is a diagonal matrix; and we have
\begin{align*}
[\bm{\mu}^t_i]_d &= \mbf{k}_{im}\mbf{K}_{mm}^{-1}\mbf{m}_d^z, \quad 1 \le d \le d_{\mbf{z}}, \\
[\bm{\Sigma}^t_i]_{dd} &= k_{ii}-\mbf{k}_{im}\mbf{K}_{mm}^{-1}[\mbf{I}-\mbf{S}^z_d\mbf{K}_{mm}^{-1}]\mbf{k}_{im}^{\mathsf{T}}, \quad 1 \le d \le d_{\mbf{z}}.
\end{align*}
In the above equations, $\mbf{m}_d^z$ and $\mbf{S}^z_d$ are the mean and covariance of the inducing variables $\mbf{u}^z_d$ shared across time.
Thereafter, the ELBO over discrete time points $\{t^0, t^1, \cdots, t^L \}$ for supervised learning is derived as
\begin{align*}
\begin{split}
\mathcal{L}_{\mathrm{diffgp}} =& \mathbb{E}_{q(\mbf{f}|\mbf{z}^L)q(\mbf{z}^L|\mbf{x})} [ \log p(\mbf{y}|\mbf{f}) ] \\
&- \mathrm{KL}[q(\mbf{u}^z)||p(\mbf{u}^z)] - \mathrm{KL}[q(\mbf{u})||p(\mbf{u})].
\end{split}
\end{align*}
Different from~\eqref{eq_elbo_sde}, the sparse GP assisted SDE here results in analytical KL terms due to the Gaussian posteriors.

\section{$\beta$-ELBO interpreted from VIB} \label{app_vib}
We can interpret the DLVKL from the view of variational information bottleneck (VIB)~\cite{alemi2016deep}. Suppose that $\mbf{z}$ is a stochastic encoding of the input $\mbf{x}$, our goal is to learn an encoding that is maximally informative about the output $\mbf{y}$ subject to a constraint on its complexity as
\begin{align*} \label{eq_vib}
\mbox{max } \mathbb{I}[\mbf{z}, \mbf{y}], \mbox{ s.t. } \mathbb{I}_q[\mbf{z}, \mbf{x}] \le I_c,
\end{align*} 
where $\mathbb{I}[.,.]$ represents the mutual information (MI) between two random variables. The above constraint is applied in order to learn a good representation rather than the simple identity $\mbf{z} = \mbf{x}$. By introducing a Lagrange multiplier $\beta$ to the above problem, we have
\begin{align*}
\mathcal{L}_{\mathrm{vib}} = \mathbb{I}[\mbf{z}, \mbf{y}] - \beta \mathbb{I}_q[\mbf{z}, \mbf{x}].
\end{align*}

As for the MI terms in $\mathcal{L}_{\mathrm{vib}}$, given the joint distribution $p(\mbf{x}, \mbf{z},\mbf{y}) = q(\mbf{z}|\mbf{x}) p_{\mathcal{D}}(\mbf{x}, \mbf{y})$, the MI $\mathbb{I}[\mbf{z}, \mbf{y}]$ is expressed as
\begin{align*}
\begin{split}
\mathbb{I}[\mbf{z}, \mbf{y}] &= \int p(\mbf{y}, \mbf{z}) \log p(\mbf{y}|\mbf{z}) d\mbf{y}d\mbf{z} + \mathbb{H}[\mbf{y}] \\
&= \mathbb{E}_{p_{\mathcal{D}}(\mbf{x}, \mbf{y})} \left[ \mathbb{E}_{q(\mbf{z}|\mbf{x})} [\log p(\mbf{y}|\mbf{z})] \right] + \mathbb{H}[\mbf{y}],
\end{split}
\end{align*}
where $p_{\mathcal{D}}(\mbf{x}, \mbf{y}) = \frac{1}{n} \sum_{i=1}^n \delta(\mbf{x} - \mbf{x}_i) \delta(\mbf{y} - \mbf{y}_i)$ is the empirical distribution estimated from training data. Note that $\mathbb{H}[\mbf{y}]$ has no trainable parameters. Besides, the MI $\mathbb{I}_q[\mbf{z}, \mbf{x}]$ is
\begin{align*}
\begin{split}
\mathbb{I}_q[\mbf{z}, \mbf{x}] &= \int q(\mbf{z}, \mbf{x}) \log \frac{q(\mbf{z}, \mbf{x})}{p(\mbf{z}) p_{\mathcal{D}}(\mbf{x})} d\mbf{z}d\mbf{x} \\
&= \mathbb{E}_{p_{\mathcal{D}}(\mbf{x})} [\mathrm{KL}(q(\mbf{z}|\mbf{x}) || p(\mbf{z}))],
\end{split}
\end{align*} 
where $p_{\mathcal{D}}(\mbf{x}) = \frac{1}{n} \sum_{i=1}^n \delta(\mbf{x} - \mbf{x}_i)$. 

Finally, we have the ELBO
\begin{align*}
\begin{split}
\mathcal{L}_{\mathrm{vib}} =& \mathbb{E}_{p_{\mathcal{D}}(\mbf{x}, \mbf{y})} \left[ \mathbb{E}_{q(\mbf{z}|\mbf{x})} [\log p(\mbf{y}|\mbf{z})] \right] \\
&- \beta \mathbb{E}_{p_{\mathcal{D}}(\mbf{x})} [\mathrm{KL}(q(\mbf{z}|\mbf{x}) || p(\mbf{z}))],
\end{split}
\end{align*}
which recovers the bound in~\eqref{eq_elbo_beta_sde} when we use sparse GP for $p(\mbf{y}|\mbf{z})$ and the NSDE transformation for $q(\mbf{z}|\mbf{x})$.

\section{Proof of Proposition~\ref{Prop_DLVKL}}
\label{app_proof}
\begin{proof}
	When the posterior collapse happens in the initial stage, we have (i) the zero KL $\mathrm{KL}[q(\mbf{z}|\mbf{x}||p(\mbf{z}|\mbf{x})]$ staying at its local minimum; and (ii) the mapped inputs $\mbf{z}_i \sim \mathcal{N}(\mbf{0}, \mbf{I})$, $0 \le i \le n$ and inducing inputs $\widetilde{\mbf{z}}_i \sim \mathcal{N}(\mbf{0}, \mbf{I})$, $0 \le i \le m$. As a result, the relative distance between any two inputs always follow $\bm{\tau} \sim \mathcal{N}(\mbf{0}, \mbf{I})$. We therefore have the collapsed kernel value
	\begin{align*}
	\mathbb{E}[k(\mbf{z}_i, \mbf{z}'_i)] = \mathbb{E}[k(\mbf{z}_i, \widetilde{\mbf{z}}'_i)] = h^2 \mathbb{E} [g_{\bm{\psi}}(\bm{\tau}_i)] = c_{\bm{\theta}},
	\end{align*}
	where $c_{\bm{\theta}}$, which is composed of the model parameters $\bm{\theta}$, is independent of inputs. This makes $\mbf{K}_{mm}$ and $\mbf{K}_{nn}$ be the matrices with all the elements being the same, which however are not invertible. In practice, we usually add a positive numeric jitter to the diagonal elements of $\mbf{K}_{mm}$ and $\mbf{K}_{nn}$ in order to relieve this issue.
	
	When we are attempting to optimize the GP parameters, the collapsed kernel cannot measure the correlations among inputs. In this case, it is observed that the posterior mean for $q(\mbf{f}_d|\mbf{x})$ follows
	\begin{align*}
	\bm{\mu}^f_d = \mbf{K}_{nm}\mbf{K}_{mm}^{-1}\mbf{m}_d \propto \overline{m}_d \mbf{1}^{\mathsf{T}},
	\end{align*}
	where $\overline{m}_d$ is the average of $\mbf{m}_d$. It indicates that the GP degenerates to a constant predictor. For example, we know that the optimum of the mean $\mbf{m}_d$ of $q(\mbf{u}_d)$ for GP regression satisfies~\cite{titsias2009variational}
	\begin{align*}
	\mbf{m}_d = \frac{1}{\nu^{\epsilon}_d} \mbf{K}_{mm} (\mbf{K}_{mm} + \frac{1}{\nu^{\epsilon}_d} \mbf{K}_{mn}\mbf{K}_{nm})^{-1} \mbf{K}_{mn} \mbf{y}_d \propto \mbf{1}_m \mbf{1}_n^{\mathsf{T}} \mbf{y}_d.
	\end{align*}
	When $\mbf{y}_d$ is normally normalized, i.e., $\mathbb{E}[\mbf{y}_d]=0$, we have $\mbf{m}_d = \mbf{0}$ and therefore $\bm{\mu}^f_d = \mbf{0}$. As for classification, the degenerated constant predictor will simply use the percentages of training labels as class probabilities.
	
	Hence, due to the constant prediction mean, what the optimizer can do is adjusting all the parameters of the encoder and GP for simply calibrating the prediction variances in order to fit the output variances in training data.
	\qedhere
\end{proof}

\section{Experimental details} \label{app_exp_details}
\textbf{Toy cases.} For the two toy cases, we adopt the settings for the following regression and classification tasks except that (i) the inducing size is $m=20$; (ii) the length-scales of the RBF kernel is initialized as 1.0; (iii) the batch size is $\mathtt{min}\{64, n\}$; and finally (iv) the Adam optimizer is ran over 5000 iterations.

\textbf{Regression and classification tasks.} The experimental configurations for the six regression tasks ($\mathtt{boston}$, $\mathtt{concrte}$, $\mathtt{wine}$-$\mathtt{red}$, $\mathtt{keggdirected}$, $\mathtt{kin40k}$, $\mathtt{protein}$) and two classification tasks ($\mathtt{connect}$-$\mathtt{4}$, $\mathtt{higgs}$) are elaborated as below. 

As for data preprocessing, we perform standardization over input dimensions to have zero mean and unit variance. Additionally, the outputs are standardized for regression. We shuffle the data and randomly select 10\% for testing. The shuffle and split are repeated to have ten instances for multiple runs.

As for the GP part, we adopt the RBF kernel with the length-scales initialized as $0.1\sqrt{d_{\mbf{z}}}$ and the signal variance initialized as $1.0$. The inducing size is $m=100$. The related positions of inducing points $\widetilde{\mbf{x}}$ are initialized in the original input space $\mathcal{X}$ through clustering techniques and then passed through the SDE transformation as $\widetilde{\mbf{z}} = \mbox{SDE}(\widetilde{\mbf{x}})$ for DLVKL-NSDE. The variational parameters for the inducing variables $\mbf{u}_d \sim \mathcal{N}(\mbf{u}_d|\mbf{m}_d, \mbf{S}_d)$ are initialized as $\mbf{m}_d = \mbf{0}$ and $\mbf{S}_d = \mbf{I}$. We set the prior parameter as $\beta = 1.0$ on the small $\mathtt{boston}$, $\mathtt{concrete}$ and $\mathtt{wine}$-$\mathtt{red}$ datasets and $\beta = 10^{-2}$ on the remaining datasets, and have the SDE parameters as $T = 1.0$ and $L = 10$. 

As for the MLP part, we use the fully-connected (FC) NN with three hidden layers and the ReLU activation. The number of units for the hidden layers is $\mathrm{max}[2 d_{\mbf{x}}, 10]$. Particularly, the MLPs of DLVKL-NSDE in~\eqref{eq_mlp_drift_diff} share the hidden layers but have separate output layers for explaining the drift and diffusion, respectively. The diffusion variance $\nu_0$ is initialized as $0.01/T$. Additionally, since the SDE flows over time, we include time $t^l$ as additional inputs for the MLPs. And all the layers except the output layers share the parameters over time.

As for the optimization, we employ the Adam with the batch size of 256, the learning rate of $5\times 10^{-3}$,\footnote{The training of GPs with Adam often uses the learning rate of $10^{-2}$. While the training of NN often uses the learning rate of $10^{-3}$. Since the DLVKL-NSDE is a hybrid model, we adopt a medium learning rate of $5\times10^{-3}$. } and the maximum number of iterations as 3000 on the small $\mathtt{boston}$, $\mathtt{concrete}$ and $\mathtt{wine}$-$\mathtt{red}$ datasets, 100000 on the large $\mathtt{higgs}$ dataset, and 20000 on the remaining datasets. In the experiments, we do not adopt additional fine-tune tricks, e.g., the scheduled learning rate, the regularized weights, or the dropout technique, for MLPs. 

\textbf{The $\mathtt{cifar}$-$\mathtt{10}$ image classification task.} For this image classification dataset, we build our codes upon the resnet-20 architecture implemented at \url{https://github.com/yxlijun/cifar-tensorflow}. We keep the convolution layers and the 64D FC layer, but additionally add the ``FC(10+1)-tanh-FC(10+1)-tanh-FC(2$\times1$0)'' layers plus the GP part for DLVKL-NSDE. For DKL, we drop the additional time input and use 10 units in the final layer. We use $m=500$ inducing points and directly initialize them in the latent space, since the encoded inducing strategy in the high-dimensional image space yields too many parameters which may make the model training difficult. We use the default data split, data augmentation and optimization strategies of resent-20 and run over 200 epochs.

\textbf{The $\mathtt{mnist}$ unsupervised learning task.} For the $\mathtt{mnist}$ dataset, the intensity of the gray images is normalized to $[0,1]$. We build the decoder for the models using FC nets, the architecture of which is ``784 Inputs-FC(196)-Relu-FC(49)-Relu-FC(2$\times$2)''. Note that the DKL employs a deterministic encoder with the last layer as FC(2). The VAE uses a mirrored NN structure to build the corresponding decoder. Differently, the DKL and DLVKL adopt the sparse GP decoder for mapping the 2D latent inputs to 784 outputs using the shared RBF kernel with the length-scales initialized as $1.0$ and the signal variance initialized as $1.0$. The inducing size is $m = 100$ and the related positions are optimized through the encoded inducing strategy. The mean for the prior $p(\mbf{z}|\mbf{x})$ is obtained through the PCA transformation of $\mbf{x}$. Finally, we employ the Adam optimizer with the batch size of 256, the learning rate of $5\times 10^{-3}$, and run it over 20000 iterations.

\ifCLASSOPTIONcaptionsoff
\newpage
\fi



%

\bibliographystyle{IEEEtran}
\bibliography{IEEEabrv,DLVKLNSDE}

\begin{thebibliography}{10}
\providecommand{\url}[1]{#1}
\csname url@samestyle\endcsname
\providecommand{\newblock}{\relax}
\providecommand{\bibinfo}[2]{#2}
\providecommand{\BIBentrySTDinterwordspacing}{\spaceskip=0pt\relax}
\providecommand{\BIBentryALTinterwordstretchfactor}{4}
\providecommand{\BIBentryALTinterwordspacing}{\spaceskip=\fontdimen2\font plus
\BIBentryALTinterwordstretchfactor\fontdimen3\font minus
  \fontdimen4\font\relax}
\providecommand{\BIBforeignlanguage}[2]{{%
\expandafter\ifx\csname l@#1\endcsname\relax
\typeout{** WARNING: IEEEtran.bst: No hyphenation pattern has been}%
\typeout{** loaded for the language `#1'. Using the pattern for}%
\typeout{** the default language instead.}%
\else
\language=\csname l@#1\endcsname
\fi
#2}}
\providecommand{\BIBdecl}{\relax}
\BIBdecl

\bibitem{williams2006gaussian}
C.~K. Williams and C.~E. Rasmussen, \emph{Gaussian processes for machine
  learning}.\hskip 1em plus 0.5em minus 0.4em\relax MIT press Cambridge, MA,
  2006.

\bibitem{wang2014spectrum}
L.~{Wang} and C.~{Li}, ``Spectrum-based kernel length estimation for {G}aussian
  process classification,'' \emph{IEEE Transactions on Cybernetics}, vol.~44,
  no.~6, pp. 805--816, 2014.

\bibitem{li2020shared}
P.~{Li} and S.~{Chen}, ``Shared {G}aussian process latent variable model for
  incomplete multiview clustering,'' \emph{IEEE Transactions on Cybernetics},
  vol.~50, no.~1, pp. 61--73, 2020.

\bibitem{lawrence2005probabilistic}
N.~Lawrence, ``Probabilistic non-linear principal component analysis with
  {G}aussian process latent variable models,'' \emph{Journal of Machine
  Learning Research}, vol.~6, no. Nov, pp. 1783--1816, 2005.

\bibitem{frigola2014variational}
R.~Frigola, Y.~Chen, and C.~E. Rasmussen, ``Variational {G}aussian process
  state-space models,'' in \emph{Advances in Neural Information Processing
  Systems}, 2014, pp. 3680--3688.

\bibitem{liu2018remarks}
H.~Liu, J.~Cai, and Y.-S. Ong, ``Remarks on multi-output {G}aussian process
  regression,'' \emph{Knowledge-Based Systems}, vol. 144, no. March, pp.
  102--121, 2018.

\bibitem{shahriari2016taking}
B.~Shahriari, K.~Swersky, Z.~Wang, R.~P. Adams, and N.~de~Freitas, ``Taking the
  human out of the loop: {A} review of {B}ayesian optimization,''
  \emph{Proceedings of the IEEE}, vol. 104, no.~1, pp. 148--175, 2016.

\bibitem{luo2019evolutionary}
J.~{Luo}, A.~{Gupta}, Y.-S. {Ong}, and Z.~{Wang}, ``Evolutionary optimization
  of expensive multiobjective problems with co-sub-pareto front {G}aussian
  process surrogates,'' \emph{IEEE Transactions on Cybernetics}, vol.~49,
  no.~5, pp. 1708--1721, 2019.

\bibitem{snelson2006sparse}
E.~Snelson and Z.~Ghahramani, ``Sparse gaussian processes using
  pseudo-inputs,'' in \emph{Advances in Neural Information Processing
  Systems}.\hskip 1em plus 0.5em minus 0.4em\relax MIT Press, 2006, pp.
  1257--1264.

\bibitem{titsias2009variational}
M.~Titsias, ``Variational learning of inducing variables in sparse {G}aussian
  processes,'' in \emph{Artificial Intelligence and Statistics}, 2009, pp.
  567--574.

\bibitem{kingma2014adam}
D.~P. Kingma and J.~Ba, ``Adam: {A} method for stochastic optimization,''
  \emph{arXiv preprint arXiv:1412.6980}, 2014.

\bibitem{hensman2013gaussian}
J.~Hensman, N.~Fusi, and N.~D. Lawrence, ``Gaussian processes for big data,''
  in \emph{Uncertainty in Artificial Intelligence}.\hskip 1em plus 0.5em minus
  0.4em\relax Citeseer, 2013, p. 282.

\bibitem{wilson2015kernel}
A.~Wilson and H.~Nickisch, ``Kernel interpolation for scalable structured
  gaussian processes (kiss-gp),'' in \emph{International Conference on Machine
  Learning}.\hskip 1em plus 0.5em minus 0.4em\relax PMLR, 2015, pp. 1775--1784.

\bibitem{gardner2018product}
J.~R. Gardner, G.~Pleiss, R.~Wu, K.~Q. Weinberger, and A.~G. Wilson, ``Product
  kernel interpolation for scalable {G}aussian processes,'' in \emph{Artificial
  Intelligence and Statistics}, 2018, pp. 1407--1416.

\bibitem{gal2014distributed}
Y.~{Gal}, M.~van~der {Wilk}, and C.~{Rasmussen}, ``Distributed variational
  inference in sparse {G}aussian process regression and latent variable
  models,'' in \emph{Advances in Neural Information Processing Systems}, 2014,
  pp. 3257--3265.

\bibitem{peng2017asynchronous}
H.~Peng, S.~Zhe, X.~Zhang, and Y.~Qi, ``Asynchronous distributed variational
  gaussian process for regression,'' in \emph{International Conference on
  Machine Learning}.\hskip 1em plus 0.5em minus 0.4em\relax PMLR, 2017, pp.
  2788--2797.

\bibitem{deisenroth2015distributed}
M.~P. Deisenroth and J.~W. Ng, ``Distributed gaussian processes,'' in
  \emph{International Conference on Machine Learning}.\hskip 1em plus 0.5em
  minus 0.4em\relax PMLR, 2015, pp. 1481--1490.

\bibitem{liu2018generalized}
H.~Liu, J.~Cai, Y.~Wang, and Y.-S. Ong, ``Generalized robust {B}ayesian
  committee machine for large-scale {G}aussian process regression,'' in
  \emph{International Conference on Machine Learning}, 2018, pp. 1--10.

\bibitem{liu2020gaussian}
H.~Liu, Y.-S. Ong, X.~Shen, and J.~Cai, ``When {G}aussian process meets big
  data: {A} review of scalable gps,'' \emph{IEEE Transactions on Neural
  Networks and Learning Systems}, pp. 1--19, 2020.

\bibitem{lee2017deep}
J.~Lee, Y.~Bahri, R.~Novak, S.~S. Schoenholz, J.~Pennington, and
  J.~Sohl-Dickstein, ``Deep neural networks as {G}aussian processes,''
  \emph{arXiv preprint arXiv:1711.00165}, 2017.

\bibitem{cho2009kernel}
Y.~Cho and L.~K. Saul, ``Kernel methods for deep learning,'' in \emph{Advances
  in Neural Information Processing Systems}, 2009, pp. 342--350.

\bibitem{hermans2012recurrent}
M.~Hermans and B.~Schrauwen, ``Recurrent kernel machines: {C}omputing with
  infinite echo state networks,'' \emph{Neural Computation}, vol.~24, no.~1,
  pp. 104--133, 2012.

\bibitem{duvenaud2014avoiding}
D.~K. {Duvenaud}, O.~{Rippel}, R.~P. {Adams}, and Z.~{Ghahramani}, ``Avoiding
  pathologies in very deep networks,'' in \emph{Artificial Intelligence and
  Statistics}, 2014, pp. 202--210.

\bibitem{neal1996bayesian}
R.~M. Neal, \emph{Bayesian Learning for Neural Networks}.\hskip 1em plus 0.5em
  minus 0.4em\relax Berlin, Heidelberg: Springer-Verlag, 1996.

\bibitem{matthews2018gaussian}
A.~G. d.~G. Matthews, M.~Rowland, J.~Hron, R.~E. Turner, and Z.~Ghahramani,
  ``Gaussian process behaviour in wide deep neural networks,'' \emph{arXiv
  preprint arXiv:1804.11271}, 2018.

\bibitem{wilson2016deep}
A.~G. Wilson, Z.~Hu, R.~Salakhutdinov, and E.~P. Xing, ``Deep kernel
  learning,'' in \emph{Artificial Intelligence and Statistics}, 2016, pp.
  370--378.

\bibitem{wilson2016stochastic}
A.~G. Wilson, Z.~Hu, R.~R. Salakhutdinov, and E.~P. Xing, ``Stochastic
  variational deep kernel learning,'' in \emph{Advances in Neural Information
  Processing Systems}, 2016, pp. 2586--2594.

\bibitem{tran2019calibrating}
G.-L. {Tran}, E.~V. {Bonilla}, J.~P. {Cunningham}, P.~{Michiardi}, and
  M.~{Filippone}, ``Calibrating deep convolutional {G}aussian processes,'' in
  \emph{Artificial Intelligence and Statistics}, 2019, pp. 1554--1563.

\bibitem{jean2018semi}
N.~{Jean}, S.~M. {Xie}, and S.~{Ermon}, ``Semi-supervised deep kernel learning:
  {R}egression with unlabeled data by minimizing predictive variance,'' in
  \emph{Advances in Neural Information Processing Systems}, 2018, pp.
  5322--5333.

\bibitem{al2017learning}
M.~Al-Shedivat, A.~G. Wilson, Y.~Saatchi, Z.~Hu, and E.~P. Xing, ``Learning
  scalable deep kernels with recurrent structure,'' \emph{Journal of Machine
  Learning Research}, vol.~18, no.~1, pp. 2850--2886, 2017.

\bibitem{damianou2013deep}
A.~Damianou and N.~Lawrence, ``Deep {G}aussian processes,'' in \emph{Artificial
  Intelligence and Statistics}, 2013, pp. 207--215.

\bibitem{salimbeni2017doubly}
H.~Salimbeni and M.~Deisenroth, ``Doubly stochastic variational inference for
  deep {G}aussian processes,'' in \emph{Advances in Neural Information
  Processing Systems}, 2017, pp. 4588--4599.

\bibitem{kim2006bayesian}
H.-C. Kim and Z.~Ghahramani, ``Bayesian gaussian process classification with
  the {EM-EP} algorithm,'' \emph{IEEE Transactions on Pattern Analysis and
  Machine Intelligence}, vol.~28, no.~12, pp. 1948--1959, 2006.

\bibitem{nickisch2008approximations}
H.~Nickisch and C.~E. Rasmussen, ``Approximations for binary {G}aussian process
  classification,'' \emph{Journal of Machine Learning Research}, vol.~9, no.
  Oct, pp. 2035--2078, 2008.

\bibitem{damianou2016variational}
A.~C. Damianou, M.~K. Titsias, and N.~D. Lawrence, ``Variational inference for
  latent variables and uncertain inputs in {G}aussian processes,''
  \emph{Journal of Machine Learning Research}, vol.~17, no.~1, pp. 1425--1486,
  2016.

\bibitem{liu2019scalable}
H.~Liu, Y.-S. Ong, Z.~Yu, J.~Cai, and X.~Shen, ``Scalable {G}aussian process
  classification with additive noise for various likelihoods,'' \emph{arXiv
  preprint arXiv:1909.06541}, 2019.

\bibitem{titsias2010bayesian}
M.~K. {Titsias} and N.~D. {Lawrence}, ``Bayesian {G}aussian process latent
  variable model,'' in \emph{International Conference on Artificial
  Intelligence and Statistics}, vol.~9, 2010, pp. 844--851.

\bibitem{kingma2013auto}
D.~P. Kingma and M.~Welling, ``Auto-encoding variational bayes,'' \emph{arXiv
  preprint arXiv:1312.6114}, 2013.

\bibitem{bowman2015generating}
S.~R. Bowman, L.~Vilnis, O.~Vinyals, A.~M. Dai, R.~Jozefowicz, and S.~Bengio,
  ``Generating sentences from a continuous space,'' \emph{arXiv preprint
  arXiv:1511.06349}, 2015.

\bibitem{bui2015stochastic}
T.~D. Bui and R.~E. Turner, ``Stochastic variational inference for {G}aussian
  process latent variable models using back constraints,'' in \emph{Black Box
  Learning and Inference NIPS workshop}, 2015.

\bibitem{friedrich2011approaching}
R.~{Friedrich}, J.~{Peinke}, M.~{Sahimi}, and M.~R.~R. {Tabar}, ``Approaching
  complexity by stochastic methods: From biological systems to turbulence,''
  \emph{Physics Reports}, vol. 506, no.~5, pp. 87--162, 2011.

\bibitem{rezende2015variational}
D.~J. Rezende and S.~Mohamed, ``Variational inference with normalizing flows,''
  in \emph{International Conference on Machine Learning}, 2015.

\bibitem{chen2018continuous}
C.~Chen, C.~Li, L.~Chen, W.~Wang, Y.~Pu, and L.~C. Duke, ``Continuous-time
  flows for efficient inference and density estimation,'' in
  \emph{International Conference on Machine Learning}, 2018, pp. 823--832.

\bibitem{chen2018neural}
R.~T.~Q. Chen, Y.~Rubanova, J.~Bettencourt, and D.~K. Duvenaud, ``Neural
  ordinary differential equations,'' in \emph{Advances in Neural Information
  Processing Systems}, S.~Bengio, H.~Wallach, H.~Larochelle, K.~Grauman,
  N.~Cesa-Bianchi, and R.~Garnett, Eds.\hskip 1em plus 0.5em minus 0.4em\relax
  Curran Associates, Inc., 2018, pp. 6571--6583.

\bibitem{liu2019neural}
X.~Liu, T.~Xiao, S.~Si, Q.~Cao, S.~Kumar, and C.-J. Hsieh, ``Neural {SDE}:
  Stabilizing neural {ODE} networks with stochastic noise,'' \emph{arXiv
  preprint arXiv:1906.02355}, 2019.

\bibitem{hegde2019deep}
P.~Hegde, M.~Heinonen, H.~L{\"a}hdesm{\"a}ki, and S.~Kaski, ``Deep learning
  with differential {G}aussian process flows,'' in \emph{Artificial
  Intelligence and Statistics}, 2019, pp. 1812--1821.

\bibitem{chen2017variational}
X.~{Chen}, D.~P. {Kingma}, T.~{Salimans}, Y.~{Duan}, P.~{Dhariwal},
  J.~{Schulman}, I.~{Sutskever}, and P.~{Abbeel}, ``Variational lossy
  autoencoder,'' in \emph{International Conference on Learning
  Representations}, 2017.

\bibitem{he2019lagging}
J.~{He}, D.~{Spokoyny}, G.~{Neubig}, and T.~{Berg-Kirkpatrick}, ``Lagging
  inference networks and posterior collapse in variational autoencoders,'' in
  \emph{International Conference on Learning Representations}, 2019.

\bibitem{oord2016conditional}
A.~van~den {Oord}, N.~{Kalchbrenner}, O.~{Vinyals}, L.~{Espeholt}, A.~{Graves},
  and K.~{Kavukcuoglu}, ``Conditional image generation with {P}ixel{CNN}
  decoders,'' in \emph{Advances in Neural Information Processing Systems},
  2016, pp. 4797--4805.

\bibitem{tomczak2018vae}
J.~Tomczak and M.~Welling, ``Vae with a vampprior,'' in \emph{Artificial
  Intelligence and Statistics}, 2018, pp. 1214--1223.

\bibitem{he2016deep}
K.~{He}, X.~{Zhang}, S.~{Ren}, and J.~{Sun}, ``Deep residual learning for image
  recognition,'' in \emph{IEEE Conference on Computer Vision and Pattern
  Recognition}, 2016, pp. 770--778.

\bibitem{hoffman2017beta}
M.~D. Hoffman, C.~Riquelme, and M.~J. Johnson, ``The $\beta$-{VAE}’s implicit
  prior,'' in \emph{Workshop on Bayesian Deep Learning, NIPS}, 2017, pp. 1--5.

\bibitem{higgins2017beta}
I.~Higgins, L.~Matthey, A.~Pal, C.~Burgess, X.~Glorot, M.~Botvinick,
  S.~Mohamed, and A.~Lerchner, ``beta-vae: Learning basic visual concepts with
  a constrained variational framework.'' in \emph{International Conference on
  Learning Representations}, 2017, pp. 1--6.

\bibitem{sonderby2016train}
C.~K. S{\o}nderby, T.~Raiko, L.~Maal{\o}e, S.~K. S{\o}nderby, and O.~Winther,
  ``How to train deep variational autoencoders and probabilistic ladder
  networks,'' in \emph{International Conference on Machine Learning}, 2016.

\bibitem{fu2019cyclical}
H.~Fu, C.~Li, X.~Liu, J.~Gao, A.~Celikyilmaz, and L.~Carin, ``Cyclical
  annealing schedule: {A} simple approach to mitigating kl vanishing,'' in
  \emph{The Association for Computational Linguistics: Human Language
  Technologies}, 2019, pp. 240--250.

\bibitem{alemi2016deep}
A.~A. Alemi, I.~Fischer, J.~V. Dillon, and K.~Murphy, ``Deep variational
  information bottleneck,'' \emph{arXiv preprint arXiv:1612.00410}, 2016.

\bibitem{van2017convolutional}
M.~van~der Wilk, C.~E. Rasmussen, and J.~Hensman, ``Convolutional {G}aussian
  processes,'' in \emph{Advances in Neural Information Processing Systems},
  2017, pp. 2845--2854.

\end{thebibliography}

\end{document}